\documentclass[lettersize,journal]{IEEEtran}

\usepackage{cite}
\usepackage{graphicx}
\usepackage{url}
\usepackage{amsmath}
\usepackage{amssymb}
\usepackage[caption = false, font=footnotesize]{subfig}
\usepackage{float}
\usepackage{booktabs} % Allows the use of \toprule, \midrule and \bottomrule in tables for horizontal lines
\usepackage{multirow}
\usepackage[colorlinks=true,linkcolor=blue]{hyperref}%
\usepackage{array}
\usepackage{color}
\usepackage{tabularx}
\usepackage{longtable}
\usepackage{tabu}
\usepackage{pifont}% http://ctan.org/pkg/pifont
\usepackage{amssymb}
\usepackage{ragged2e}
\usepackage{xcolor}
\usepackage{colortbl, booktabs}
\usepackage{algorithm}  
\usepackage{algpseudocode}

\begin{document}

\title{Standardizing Generative Face Video Compression using Supplemental Enhancement Information}

\author{Bolin Chen,~\IEEEmembership{Member,~IEEE}, Yan Ye,~\IEEEmembership{Senior Member,~IEEE}, Jie Chen, Ru-Ling Liao, Shanzhi Yin, Shiqi Wang,~\IEEEmembership{Senior Member,~IEEE}, Kaifa Yang, Yue Li, Yiling Xu,~\IEEEmembership{Member,~IEEE}, Ye-Kui Wang, Shiv Gehlot, Guan-Ming Su,~\IEEEmembership{Senior Member,~IEEE}, Peng Yin, Sean McCarthy, Gary J. Sullivan,~\IEEEmembership{Fellow,~IEEE}
\IEEEcompsocitemizethanks{
\IEEEcompsocthanksitem Bolin Chen, Yan Ye, Jie Chen and Ru-Ling Liao are with DAMO Academy, Alibaba Group and Hupan Lab (E-mail: \{chenbolin.chenboli, yan.ye, jiechen.cj, ruling.lrl\}@alibaba-inc.com).\\
\IEEEcompsocthanksitem Shanzhi Yin and Shiqi Wang are with the Department of Computer Science, City University of Hong Kong (E-mail: shanzhyin3-c@my.cityu.edu.hk, shiqwang@cityu.edu.hk). \\
\IEEEcompsocthanksitem Kaifa Yang, Yue Li and Yiling Xu are with School of Electronic Information and Electronic Engineering, Shanghai Jiao Tong University (E-mail: \{sekiroyyy, yueli983, yl.xu\}@sjtu.edu.cn).\\
\IEEEcompsocthanksitem Ye-Kui Wang is with Bytedance Inc. (E-mail: yekui.wang@bytedance.com).\\
\IEEEcompsocthanksitem Shiv Gehlot, Guan-Ming Su, Peng Yin, Sean McCarthy and Gary J. Sullivan are with Dolby Laboratories, Inc. (E-mail: \{shiv.gehlot, guanming.su, pyin, sean.mccarthy, gary.sullivan\}@dolby.com).}
}

\IEEEtitleabstractindextext{%
\begin{abstract}
\justifying
This paper proposes a Generative Face Video Compression (GFVC) approach using Supplemental Enhancement Information (SEI), where a series of compact spatial and temporal representations of a face video signal (e.g., 2D/3D key-points, facial semantics and compact features) can be coded using SEI messages and inserted into the coded video bitstream. At the time of writing, the proposed GFVC approach using SEI messages has been included into a draft amendment of the Versatile Supplemental Enhancement Information (VSEI) standard by the Joint Video Experts Team (JVET) of ISO/IEC JTC 1/SC 29 and ITU-T SG21, which will be standardized as a new version of ITU-T H.274 $|$ ISO/IEC 23002-7.
To the best of the authors’ knowledge, the JVET work on the proposed SEI-based GFVC approach is the first standardization activity for generative video compression. The proposed SEI approach has not only advanced the reconstruction quality of early-day Model-Based Coding (MBC) via the state-of-the-art generative technique, but also established a new SEI definition for future GFVC applications and deployment. Experimental results illustrate that the proposed SEI-based GFVC approach can achieve remarkable rate-distortion performance compared with the latest Versatile Video Coding (VVC) standard, whilst also potentially enabling a wide variety of functionalities including user-specified animation/filtering and metaverse-related applications. 
\end{abstract}
}

\maketitle
\IEEEdisplaynontitleabstractindextext

% \fi
\IEEEpeerreviewmaketitle

% \ifCLASSOPTIONcompsoc
% \IEEEraisesectionheading{\section{Introduction}\label{sec:introduction}}
% \else
% \section{Introduction}
% \label{sec:introduction}
% \fi


 

% \vspace{-1mm}
\section{Introduction}
\begin{figure*}[tb]
\vspace{-1em}
\centering
\centerline{\includegraphics[width=1 \textwidth]{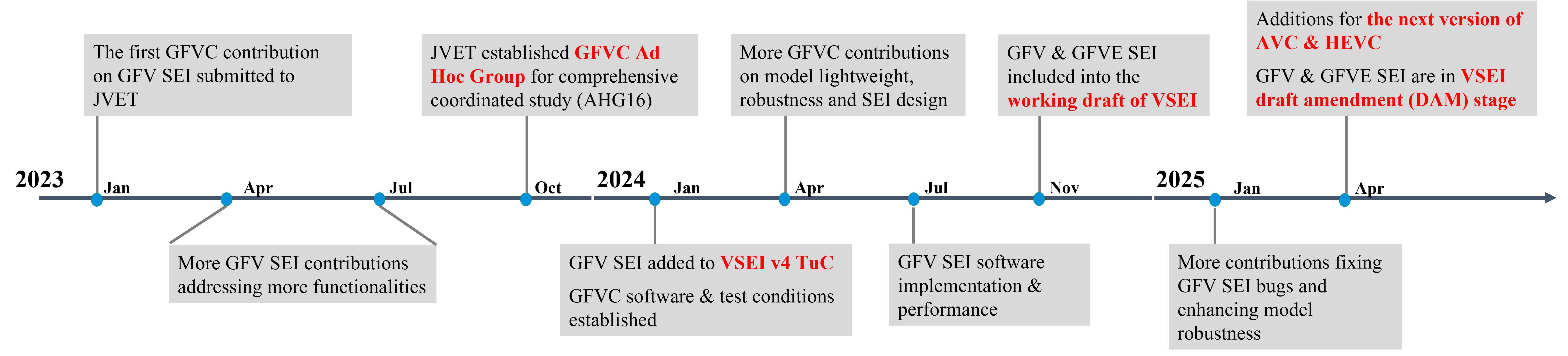}}  
\caption{Roadmap of JVET GFVC standardization activities over the past two and a half years.}
% \vspace{-1em}
\label{fig_roadmap} 
\end{figure*}

% \vspace{-1mm}
\IEEEPARstart{R}{ecently}, generative AI technology has greatly accelerated the progress of Generative Face Video Compression (GFVC)~\cite{chen2023generative,ultralow,10743340,Chen2025DCC,wang2021Nvidia,CHEN2022DCC}, showing promising performance potentials and application diversities over traditional codecs~\cite{wiegand2003overview,sullivan2012overview,bross2021overview}. Different from early-day Model-Based Coding (MBC) techniques~\cite{7268565,1989Object,1457470,lopez1995head,150969}, GFVC exploits the strong inference capabilities of deep generative models to improve the face reconstruction quality using very compact spatial and temporal representations (e.g., 2D/3D keypoints, facial semantics and compact features). Inspired by promising rate-distortion performance and the potential to enable more diverse applications, the Joint Video Experts Team (JVET) of ISO/IEC JTC 1/SC 29 and ITU-T SG21 (formerly SG16) has established an Ad hoc Group (AhG) for comprehensive GFVC investigations, including coordinated experimentation, software implementation, interoperability study, and so on. As illustrated in Fig. \ref{fig_roadmap}, over the past two and a half years, the standardization efforts of GFVC have made significant strides towards establishing a unified test environment and bitstream syntax structure for coding of face video content.
In particular, the JVET GFVC AhG has established GFVC test conditions~\cite{JVET-AJ2035} and software tools~\cite{JVET-AG0042,JVET-AH0114}, and the Generative Face Video (GFV) and Generative Face Video Enhancement (GFVE) Supplemental Enhancement Information (SEI) messages have been included into the Versatile Supplemental Enhancement Information (VSEI) standard~\cite{JVET-AJ2006}, which will be  standardized as a new version of ITU-T H.274 $|$ ISO/IEC 23002-7. Also, the GFV and GFVE SEI messages have been recently added to the next version~\cite{JVET-AL0148} of H.264/Advanced Video Coding (AVC)~\cite{wiegand2003overview} and H.265/High Efficiency Video Coding (HEVC)~\cite{sullivan2012overview}. 

From the perspective of JVET standardization activities, the proposed approach provides a reasonable solution to a standardized GFVC approach using SEI messages, bridging deep generative models with face video compression tasks using a standardized bitstream format. In particular, standardizing GFVC with SEI messages allows various GFVC feature formats to be carried and combines these features seamlessly with the bitstreams from a traditional hybrid video codec (e.g., VVC, HEVC and AVC) without introducing normative changes to the encoder/decoder of the hybrid video codec. As a result, the proposed SEI-based GFVC standardization approach can effectively utilize advanced generative AI techniques to provide performance gains and functionalities over traditional codecs while retaining a form of backward compability with existing decoders. The main contributions of this paper can be summarized as follows,

\begin{itemize}
% \vspace{-2.5mm}
\item{The potential to standardize generative video compression with SEI messages has been explored in JVET. Taking face video coding as use case, a novel SEI-based standardization approach is proposed to specify compact facial representations using a unified high-level syntax structure that can be inserted into the coded video bitstream of a hybrid video codec.}
\item{The proposed SEI messages for generative face video compression can well support different facial representations, such as 2D/3D landmarks, 2D/3D keypoints, compact feature and facial semantics. Moreover, the GFV and GFVE SEI messages can also provide flexibility for different functionalities such as ultra-low bitrate communication, user-specified animation/filtering, and metaverse-related applications.}
\item{The proposed GFV and GFVE SEI messages have been implemented on top of the VVC reference software (VTM) and extended with common GFVC software tools, which were both endorsed by JVET as GFVC AhG activities. As such, the entire SEI-based GFVC approach can be executed in a streamlined manner. The simulation results demonstrate the effectiveness of the proposed SEI-based GFVC approach, which can achieve promising compression performance and diverse functionalities over the VVC standard. Further, upon inclusion into the next version of the HEVC and AVC standards, the proposed SEI messages have been implemented on top of the HEVC and AVC standard reference software as well.}
\end{itemize}

\section{Related Work}

\subsection{Generative Face Video Compression}
In terms of ultra-low bitrate visual compression, parameterized face video compression can be tracked to early MBC techniques~\cite{7268565,1989Object,1457470,lopez1995head,150969} in the 1990s, which provided promising compression capability via statistical priors and image synthesis. However, limited by the model learning and generalization abilities at that time, the reconstruction quality was not sufficient for the early-day MBC techniques to be further developed and applied in the following years. Recently, inspired by the strong inference capabilities of deep generative models like Generative Adversarial Networks (GANs)~\cite{goodfellow2014generative} and Diffusion Models (DMs)~\cite{NEURIPS2021_49ad23d1}, the MBC concept has been revived and further extended into a novel generative compression paradigm. Taking face video as an example, deep generative models can facilitate the characterization of a high-dimensional face signal into compact features at the encoder side, and use these features to perform accurate motion estimation for high-quality face video reconstruction at the decoder side.

In particular, keypoint-based GFVC frameworks~\cite{ultralow} can realize ultra-low bit-rate face video communication by compressing a series of learned 2D keypoints along with their local affine transformations. A wide variety of facial representations like 2D facial landmarks~\cite{9455985}, 3D keypoints~\cite{wang2021Nvidia,10811831}, compact temporal features~\cite{CHEN2022DCC}, interactive facial semantics~\cite{chen2023interactive}, progressive visual tokens~\cite{Chen2025DCC} and scalable features~\cite{chen2025scalable} have been proposed to support more robust and versatile GFVC algorithms. However, these algorithms still simply bridge generative animation models with compression tasks, and lack more mature consideration such as how to achieve commercial deployment and enable real applications through standardization. 
As for the requirements of GFVC standardization, it should focus on specifying the bitstream format for a variety of GFVC feature representations in a unified form and ensuring the interoperability/robustness of GFVC systems. Moreover, standardized GFVC format can realize ultra-low bitrate face video communication and improve the quality of experience of face video applications like video conferencing/chat. It also potentially enables a wide variety of functionalities including user-specified animation/filtering and metaverse-related applications.

\subsection{Supplemental Enhancement Information}
As additional data specified into the bitstream, SEI messages can convey extra information and support various post-processing technologies. In the past decades, JVET has successfully developed many high-level syntax features ~\cite{6324417,9395142} to enable versatile scenarios and new functionalities like viewport-adaptive 360° immersive media  and film grain removal and synthesis.

From the year of 2023, JVET has started to investigate the standardization of GFVC, covering topics from how to carry various GFVC features using standardized syntax to how to insert GFV SEI messages into the VVC coded bitstream. Specifically, the first GFV SEI message, as described in~\cite{JVET-AC0088}, was introduced to enable VVC-coded pictures to serve as base pictures, supplementing them with a minor additional bit overhead to convey facial semantics. Afterwards, a common GFV SEI message syntax~\cite{JVET-AD0051,JVET-AI0191} was proposed to support the carriage of different GFVC feature formats within a VVC coded bitstream, whilst the interfaces to GFVC translator, generator, and enhancer neural networks~\cite{JVET-AE0280,JVET-AG0088,JVET-AI0189} were also specified. 
On top of the GFV SEI message,~\cite{JVET-AH0127} proposed a new Generative Face Video Enhancement (GFVE) SEI message to support the scalable representation and layered reconstruction of GFVC. 
In addition, there were a series of syntax refinements~\cite{JVET-AI0156,JVET-AK0239,JVET-AI0184,JVET-AI0191,JVET-AL0155} and additional operations (e.g., chroma key information~\cite{JVET-AI0194}, timing information and order~\cite{JVET-AI0192,JVET-AI0186,JVET-AK0124}, low-confidence facial motion information~\cite{JVET-AF0146}, no-display flag~\cite{JVET-AI0193} and pupil position redirection~\cite{JVET-AI0137}) made available to GFV and GFVE SEI messages, which have improved their syntax accuracy and enriched their functionalities.

Moreover, interoperability among different GFVC feature formats~\cite{JVET-AG0048}, fusion/enhancement modules for quality improvement~\cite{JVET-AH0110,JVET-AH0118} and model complexity reduction for practical deployment~\cite{JVET-AG0139,JVET-AH0109} were further investigated to enable GFVC applications towards better performance. GFVC software tools~\cite{JVET-AG0042,JVET-AH0114} have been developed to support various GFVC algorithms, and the GFV/GFVE SEI messages have also been included into the working draft~\cite{JVET-AJ2006} of the VSEI standard, which will be standardized as a new version of ITU-T H.274 $|$ ISO/IEC 23002-7. Recently, the GFV and GFVE SEI messages have also been added to the next version~\cite{JVET-AL0148} of H.264/AVC~\cite{wiegand2003overview} and H.265/HEVC~\cite{sullivan2012overview}, which allows AVC- and HEVC-coded base pictures to be used in the GFVC framework.
Herein, this paper summarizes how to enable different GFVC representations with GFV SEI message. The paper also describes a technical implementation to link GFVC software tools with GFV SEI message implementation for comprehensive performance evaluations.

\begin{figure*}[tb]
\vspace{-1.8em}
\centering
\includegraphics[width=1\textwidth]{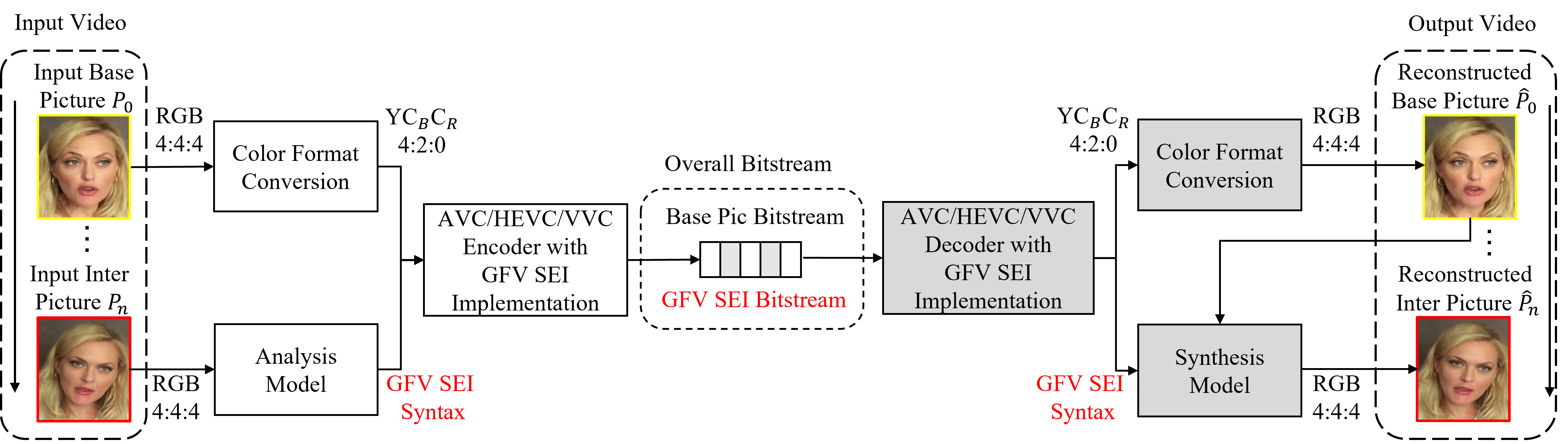} %
\caption{The encoding/decoding processes of the proposed SEI-based GFVC approach.} 
\label{GFV_approach}
\end{figure*}

\setlength{\floatsep}{1pt} % 算法之后的间隔
\setlength{\textfloatsep}{3pt} % 算法之前的间隔
\begin{algorithm}[t]
%\scriptsize
% \footnotesize
\small
%\normalsize
  \caption{Pseudocode of GFV SEI Message}
  \label{GFV_SEI_syntax}
  \begin{algorithmic}[1]
    \Require
      $\mathbf{P_{C}}$: uncompressed coordinate parameters from compact representations;
      $\mathbf{P_{M}}$: uncompressed matrix parameters from compact representations;
    \Ensure
      a standardized GFV SEI message;
       \State set $ TranslatorNN\left (\right )$, specify a neural network that may be used to convert various formats of facial parameters signaled in the SEI message into a fixed format of parameters;
       \State set $ GenerativeNN\left (\right )$, specify a neural network that may be used to generate output pictures using the fixed format of facial parameters and previously decoded output  pictures;    
       \State set ${gfv\_coordinate\_present\_flag}$, indicate whether  $\mathbf{P_{C}}$ is present;
       \State set ${gfv\_matrix\_present\_flag}$, indicate whether  $\mathbf{P_{M}}$ is present;
       \State set ${gfv\_prediction\_flag}$, determine whether to use inter-prediction for $\mathbf{P_{C}}$ and $\mathbf{P_{M}}$;
       \State set ${gfv\_precision\_factor}$, indicate the length, in bits, of syntax elements for $\mathbf{P_{C}}$ and $\mathbf{P_{M}}$;
       
      \If {${gfv\_coordinate\_present\_flag}$} 
          \If {${gfv\_prediction\_flag}$ }\State Code the difference of $\mathbf{P_{C}}$ with ${gfv\_precision\_factor}$;
          \Else \State Code $\mathbf{P_{C}}$ with ${gfv\_precision\_factor}$;
          \EndIf
      \EndIf

      \If {${gfv\_matrix\_present\_flag}$} 
          \If {${gfv\_prediction\_flag}$ }\State Code the difference of $\mathbf{P_{M}}$ with ${gfv\_precision\_factor}$;
          \Else \State Code $\mathbf{P_{M}}$ with ${gfv\_precision\_factor}$;
          \EndIf
      \EndIf
  \end{algorithmic}
\end{algorithm}

% \vspace{3cm}
\section{SEI-based Generative Compression Approach}
In this section, we provide the GFV SEI message definition and its workflow in pseudo-code format regarding how to encapsulate different GFVC feature formats with a unified GFV SEI message syntax. We also discuss how to support interoperability among different GFVC feature formats and improve reconstruction robustness via fusion/enhancement techniques. Finally, we provide implementation details of GenerativeNN$\left (  \right )$/TranslatorNN$\left (  \right )$, and describe the detailed encoding and decoding processes for the proposed SEI-based GFVC approach.

\begin{table}[t]
% \vspace{-2.3em}
\centering
\caption{Summary of compact facial representations for typical GFVC algorithms}  
\renewcommand\arraystretch{1.45}
\resizebox{1\linewidth}{!}{
\label{GFVC_representations}
\centering
\large
\begin{tabular}{c|c|c}
\hline
    GFVC Methods & Facial Representation & Parameter Format  \\ \hline
    VSBNet~\cite{9455985} & 2D Landmarks & Coordinate  \\ 
    FOMM~\cite{FOMM} & 2D Keypoints+Affine Transformation Matrices & Coordinate \& Matrix  \\ 
    DAC~\cite{ultralow} & 2D Keypoints & Coordinate  \\ 
    FV2V~\cite{wang2021Nvidia} & 3D Keypoints+Head Rotation/Translation Matrices & Coordinate \& Matrix  \\ 
    Mob M-SPADE~\cite{oquab2021low} & Segmentation Map & Matrix \\
    SNRVC~\cite{9810784} & Facial Semantics & Matrix  \\
    CFTE~\cite{CHEN2022DCC} & Compact Feature Matrices & Matrix  \\
    CTTR~\cite{chen2023csvt} & Compact Feature Matrices & Matrix  \\
    Bi-Net~\cite{9859867} & 2D Keypoints & Coordinate \\
    RDAC~\cite{konuko2023predictive} & 2D Keypoints + Residual Map & Coordinate \& Matrix  \\ 
    IFVC~\cite{chen2023interactive} & Facial Semantics  & Matrix \\ 
    PFVC~\cite{Chen2025DCC} & Progressive Tokens  & Matrix \\   \hline
        % IFVC~\cite{chen2023interactive} & Facial Semantics (Mouth/Eye/Head Rotation/Translation Matrices)  & Matrix \\ \hline
\end{tabular}
}
% \vspace{-4.5mm}
\end{table}

% \vspace{-1.2em}
\subsection{GFV SEI Message Design}
A face signal exhibits strong statistical regularities, which can be efficiently  characterized with 2D landmarks, 2D keypoints, 3D keypoints, compact features, facial semantics or  other formats. Such compact facial description strategies can lead to reduced coding bit-rate and improved coding efficiency for GFVC algorithms. However, diverse facial representations also make it difficult to standardize GFVC. Herein, we summarize a few typical GFVC algorithms and attempt to discover the characteristics of their compact representations. As shown in Table \ref{GFVC_representations}, these compact facial representations can be classified into two categories: coordinate-related parameters and matrix-related parameters. In addition, all GFVC algorithms proposed in the literature use a hybrid codec to code the base pictures. These observations confirm that there is great potential to standardize GFVC. Intuitively, a SEI message provides a reasonable standardized way to accommodate different GFVC representations into a unified format and convey these compact  representations together with VVC-coded (or HEVC- or AVC-coded) base pictures. As such, SEI-based GFVC approach has recently been a subject active study in JVET.

Algorithm~\ref{GFV_SEI_syntax} provides a pseudocode representing the syntax structure of GFV SEI message according to the GFV SEI message text that has been included in the working draft of VSEI version 4~\cite{JVET-AJ2006}. In particular, the GFV SEI message indicates facial parameters and specifies a facial parameter translator network, denoted as TranslatorNN$\left (  \right ) $, and a face picture generator neural network, denoted as GenerativeNN$\left (  \right ) $. Besides, the GFV SEI message provides specific flags/factors to determine parameter type, whether prediction is used, and codeword length. It should be emphasized that the GFV SEI message in Algorithm~\ref{GFV_SEI_syntax} is just a brief summary of the GFV SEI message included in the working draft of VSEI version 4~\cite{JVET-AJ2006}. Readers are referred to the detailed syntax and semantics of GFV SEI message draft text in~\cite{JVET-AJ2006} for further information, including 2D/3D parameters, number of coordinates, matrix dimensions and flags for extra functionalities, as well as the usage interface for TranslatorNN$\left (  \right ) $ and GenerativeNN$\left (  \right ) $ that use these GFV parameters for signal reconstruction.

\begin{figure}[tb]
\centering
% \vspace{1.2em}
\subfloat[GenerativeNN$\left (  \right )$]{\includegraphics[width=0.48\textwidth]{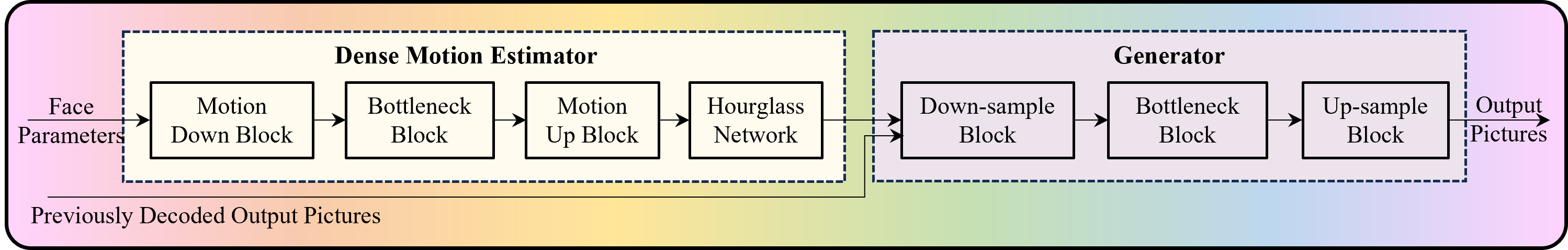}}\\
% \hspace{0.01\textwidth}
\subfloat[TranslatorNN$\left (  \right )$]{\includegraphics[width=0.48\textwidth]{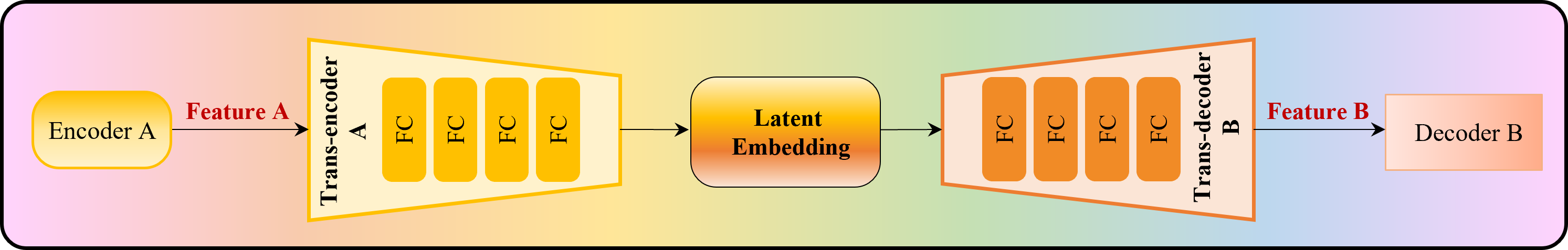}}
\caption{Illustrations of GenerativeNN$\left (  \right )$ and TranslatorNN$\left (  \right )$ functionalities for picture generation and parameter translation.} 
\label{NN_functionalities} 
\vspace{0.5em}
\end{figure}

\begin{figure*}[tb]
\centering
\vspace{-1.8em}
\subfloat[Class A]{\includegraphics[width=0.3 \textwidth]{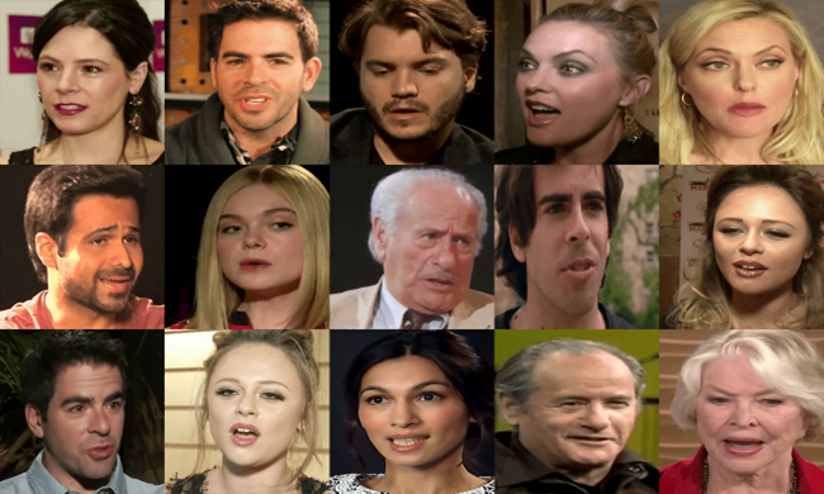}} 
\hspace{1mm}
\subfloat[Class B \& Class D]{\includegraphics[width=0.3625\textwidth]{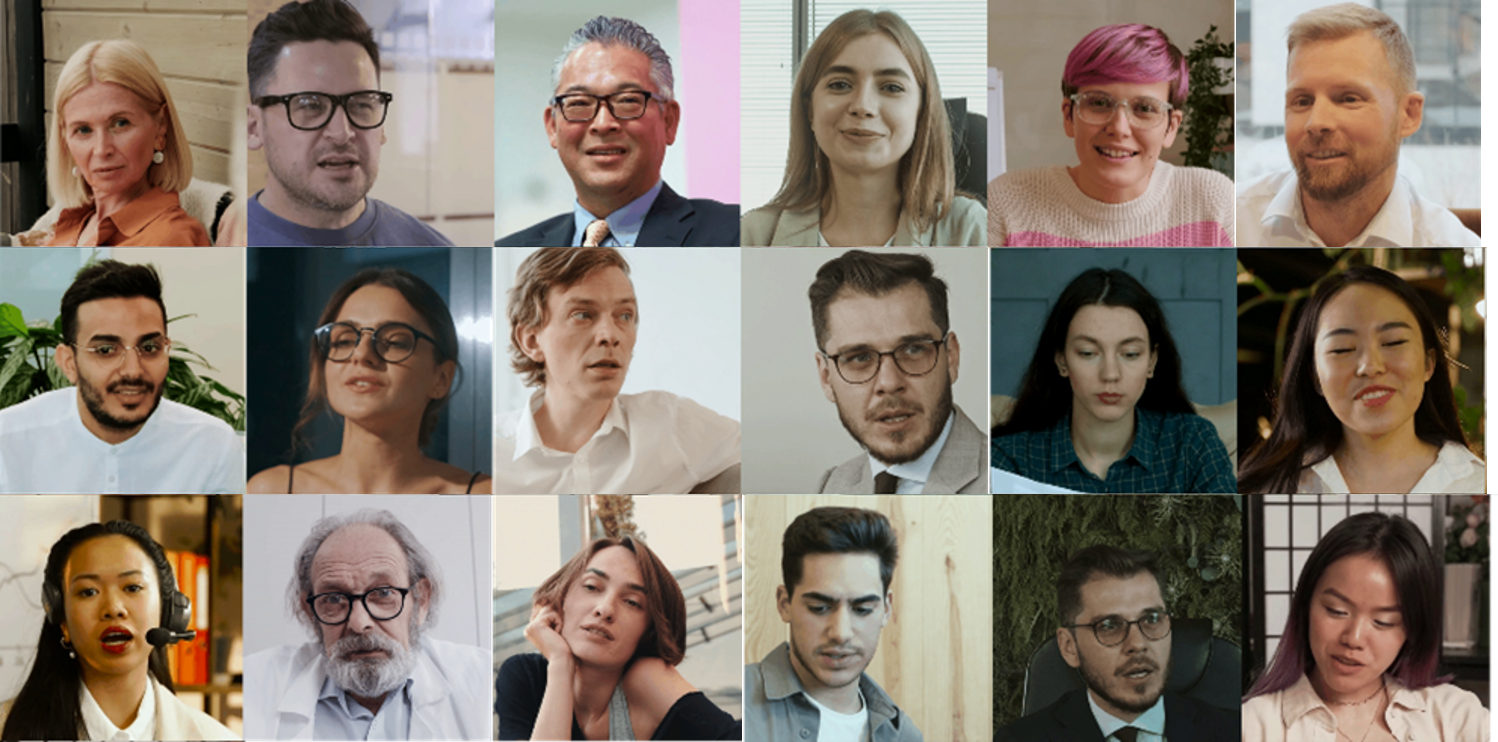}}
\hspace{1mm}
\subfloat[Class C]{\includegraphics[width=0.3\textwidth]{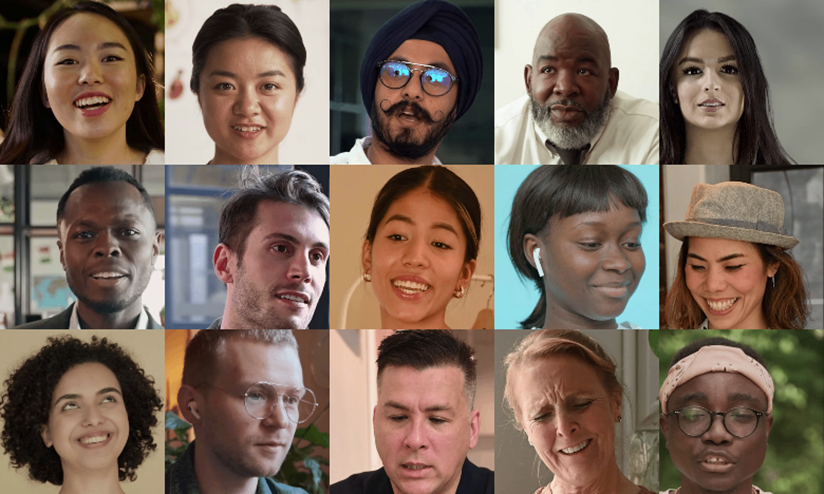}}
\caption{Common test datasets from GFVC AhG test conditions~\cite{JVET-AJ2035}. } % 
\label{CTC}
\end{figure*}

\begin{figure*}[!tb]
\centering
\vspace{-1.5em}
\subfloat[256$\times$256: Rate-DISTS]{\includegraphics[width=0.325\textwidth]{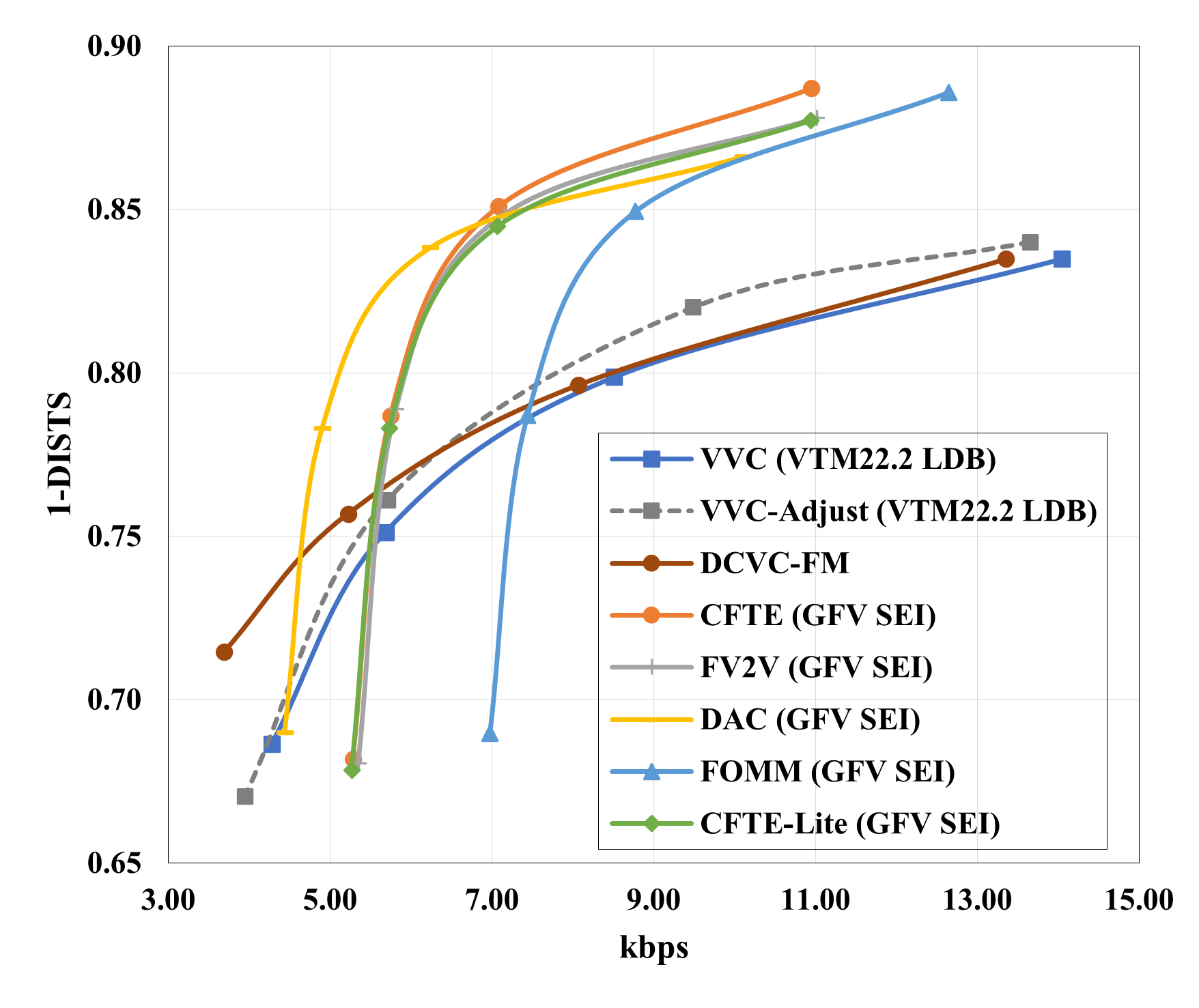}}
\subfloat[256$\times$256: Rate-LPIPS]{\includegraphics[width=0.325\textwidth,]{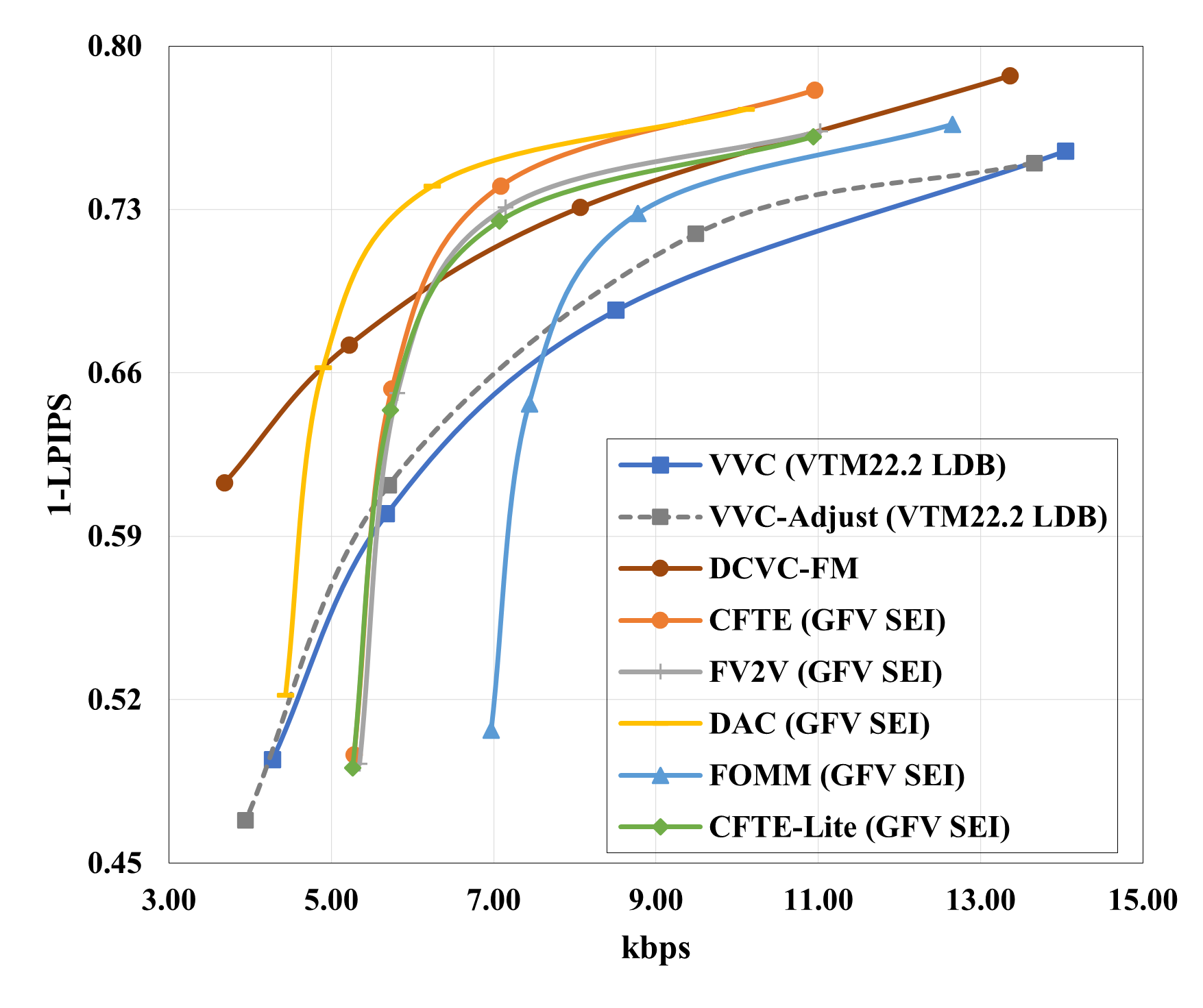}}
\subfloat[256$\times$256: Rate-PSNR]{\includegraphics[width=0.325\textwidth,]{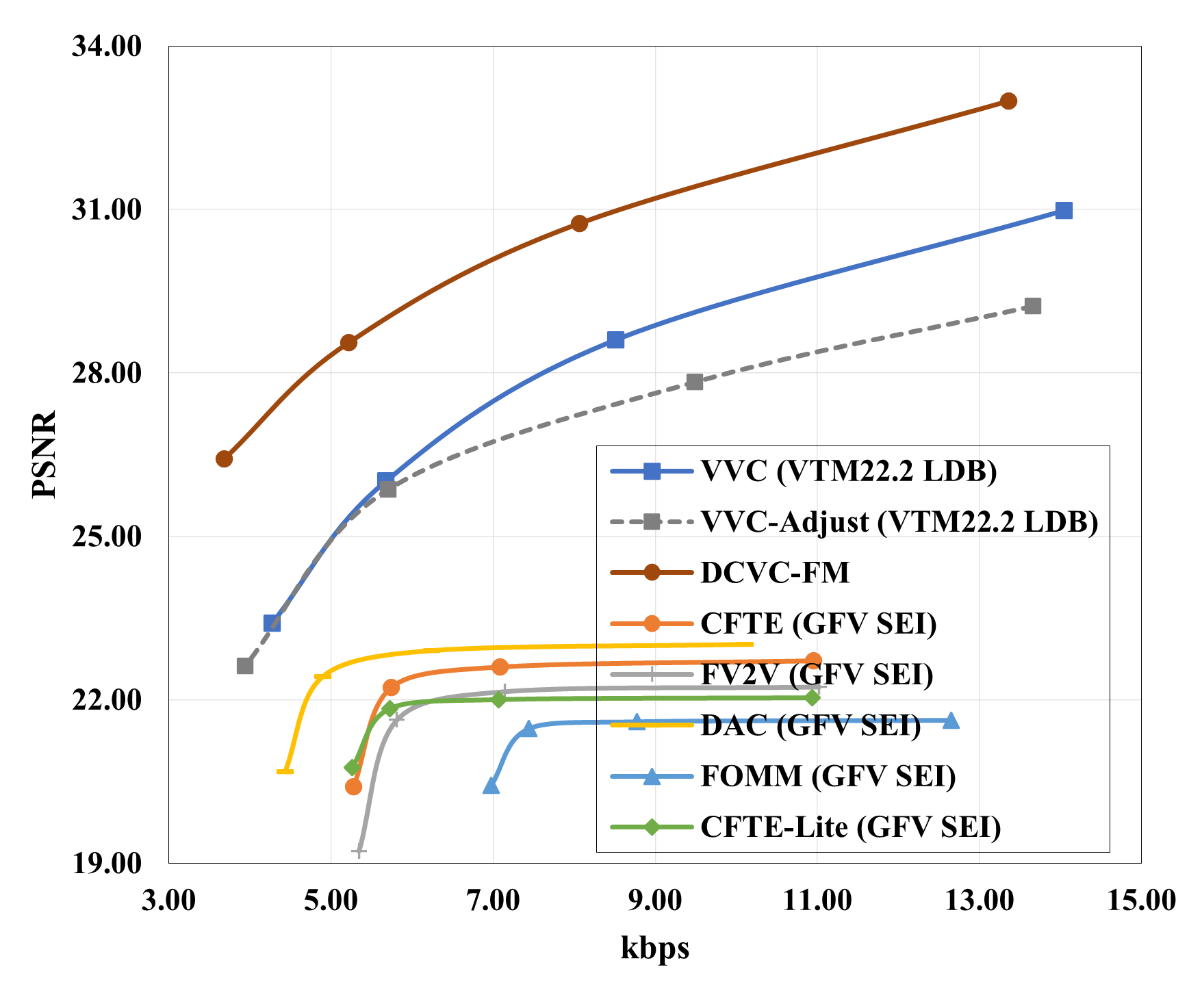}}
\\
\vspace{-0.5em}
\subfloat[512$\times$512: Rate-DISTS]{\includegraphics[width=0.325\textwidth]{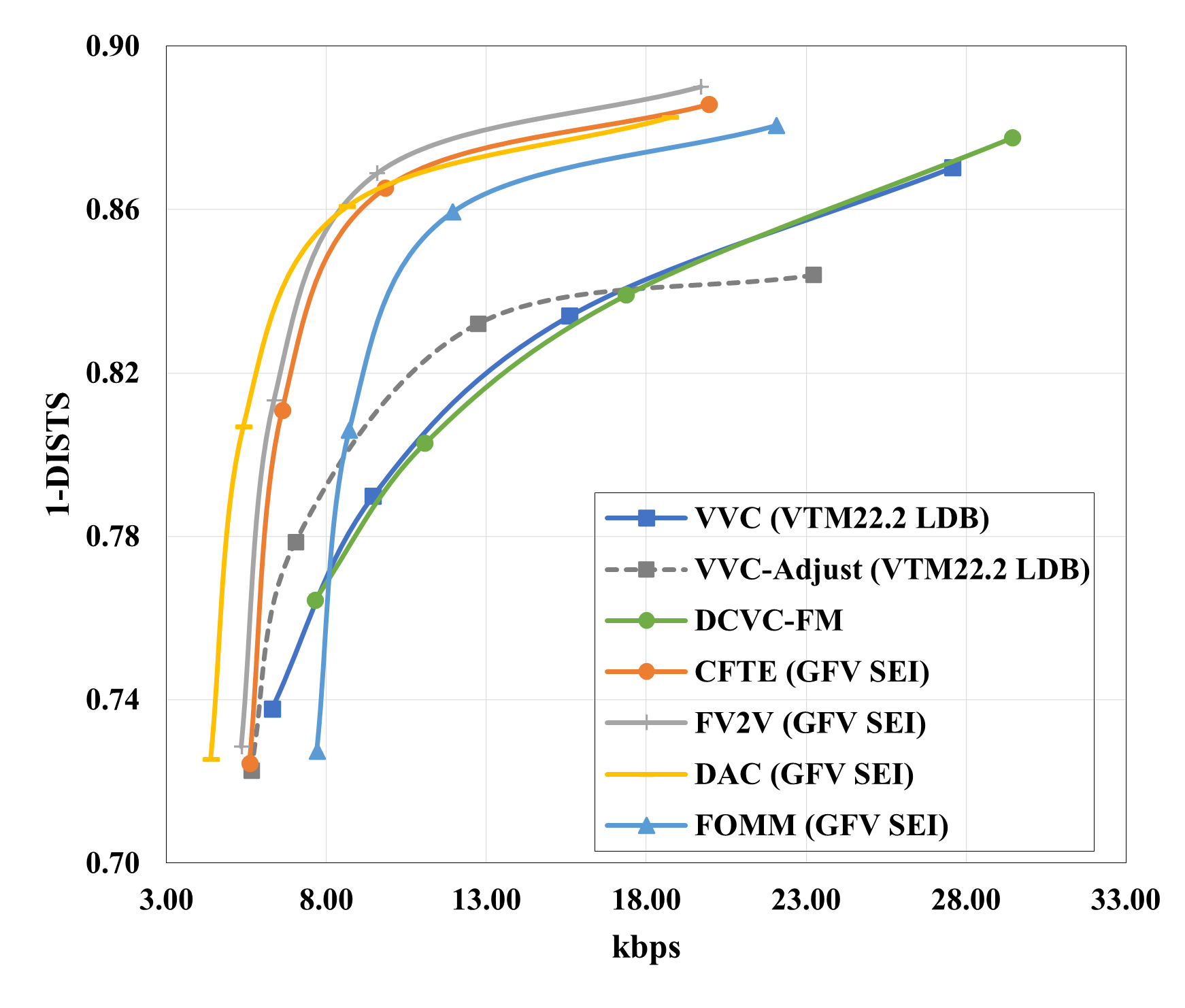}}
\subfloat[512$\times$512: Rate-LPIPS]{\includegraphics[width=0.325\textwidth]{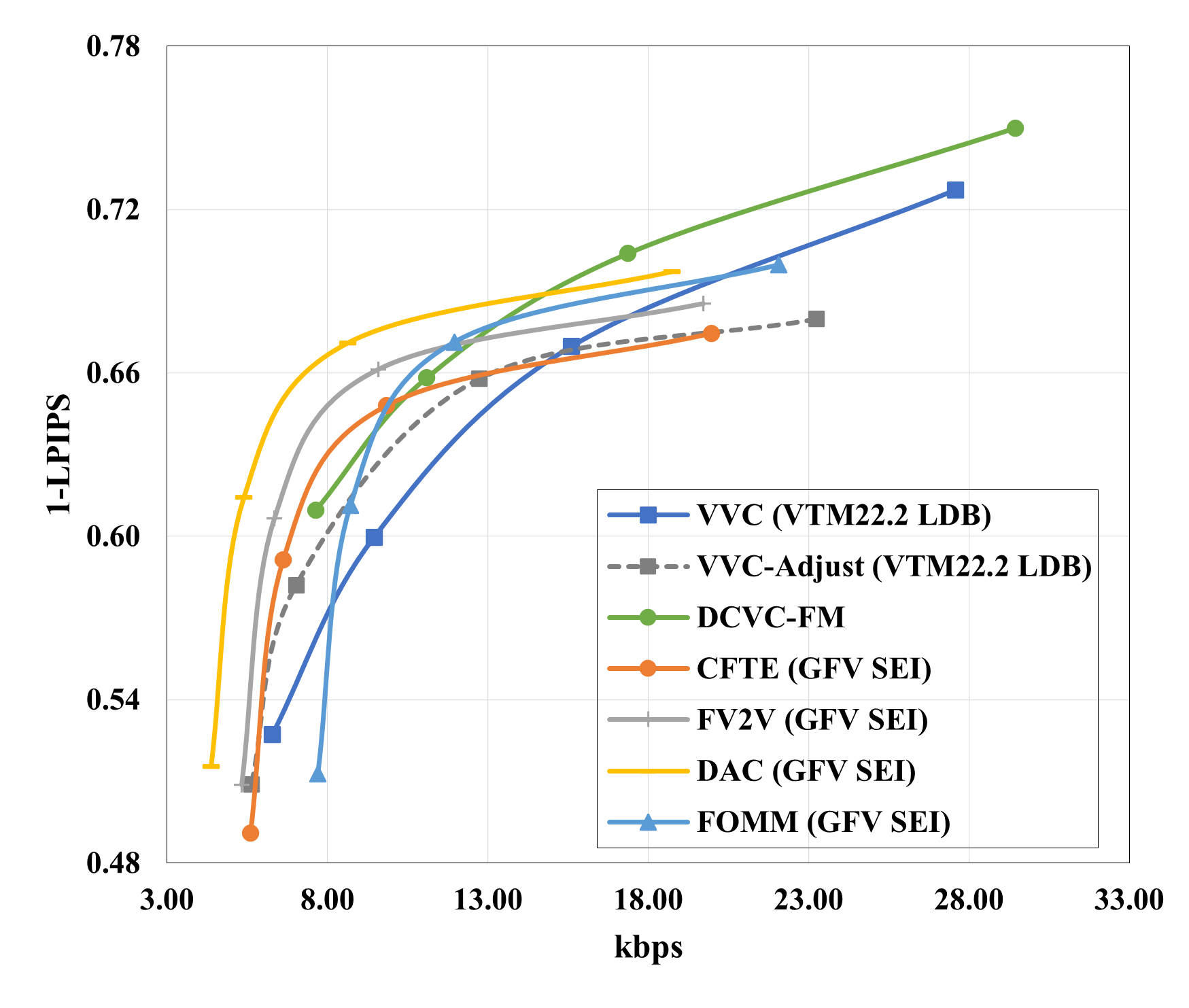}}
\subfloat[512$\times$512: Rate-PSNR]{\includegraphics[width=0.325\textwidth]{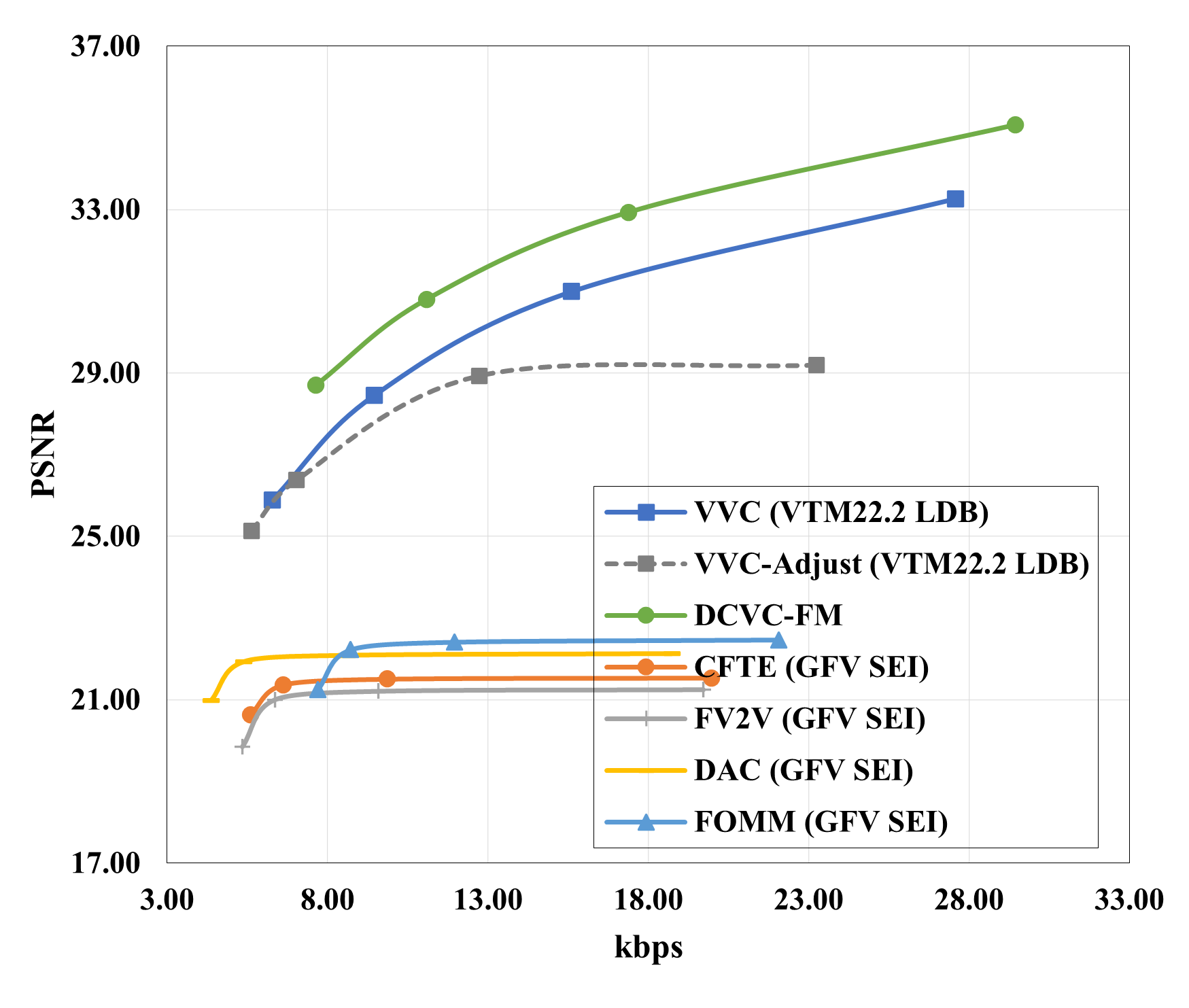}}
\caption{Rate-Distortion performance comparisons of VVC/VVC-adjust, DCVC-FM and five SEI-based GFVC approaches (CFTE, FV2V, DAC, FOMM and CFTE-Lite) in terms of DISTS, LPIPS and PSNR for both 256$\times$256 and 512$\times$512 resolutions.} % 
\label{RD}
\vspace{-1.2em}
\end{figure*}

\begin{figure*}[tb]
\centering
\vspace{-1.6em}
\subfloat[256$\times$256 Resolution]{\includegraphics[width=0.95 \textwidth]{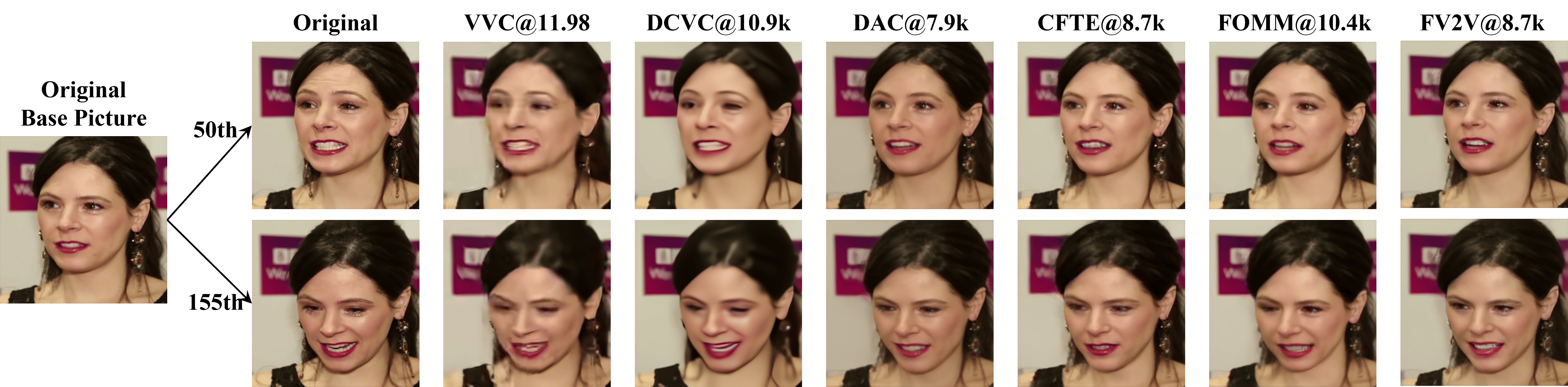}}\\
% \hspace{3em}
\subfloat[512$\times$512 Resolution]{\includegraphics[width=0.95\textwidth]{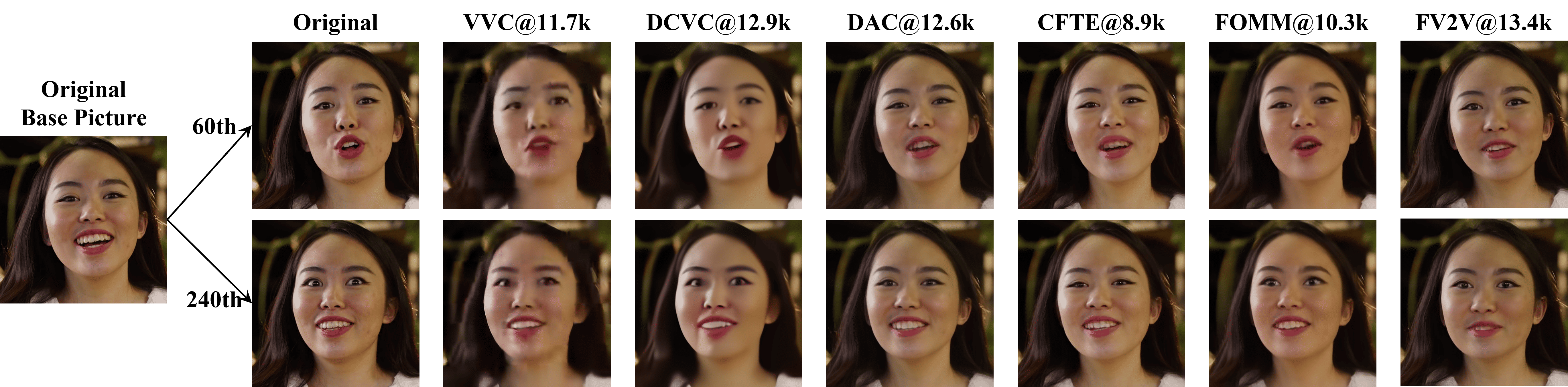}}
\caption{Subjective comparisons of VVC, DCVC-FM and four SEI-based GFVC approaches (DAC, CFTE, FOMM and FV2V) at similar bitrate.} % 
\label{subjective} 
% \vspace{-1.2em}
\end{figure*}

% \vspace{-1.2em}
\subsection{GFVC System Optimization}
The proposed GFV SEI message semantics for the VSEI standard provides a specification of syntax and semantics sufficient for usage, but does not establish conformance requirements for exactly how a decoder would use the provided SEI message to produce the pictures it would display in response to receiving the SEI message. Instead, a core decoding process is applied as for ordinary picture decoding to generate the base pictures, and the GFV processing is applied as a post-process that is outside of the motion-compensated prediction loop. This approach offers decoder implementers the freedom to customize their implementations to suit their own needs or for specifications establishing such details to be developed separately in other organizations. To facilitate this approach, the draft syntax for the GFV SEI message supports the ability for an encoder to use Uniform Resource Identifiers (URIs) to identify such third-party specifications.

Although the proposed GFV SEI message can encapsulate various GFVC features into a unified syntax format, the existence of various GFVC algorithms can still result in mismatched encoder and decoder, where the encoder chooses to send a type of facial representation that is  different from what the generator network at the decoder expects to use. In other words, various types of GFVC features with distinct characteristics may not be mutually compatible during the face generation process. To solve the issue of interoperability among different GFVC systems, JVET has included a feature translator network~\cite{JVET-AG0048,JVET-AL0147} into the usage model for the GFV SEI message. The translator network can convert received GFVC features into ones that are specific to the decoder’s preferred model for interpretation. As such, different GFVC features can be effectively adapted to one another while retaining promising coding performance.

% unchanged textures provided by a base picture and sometimes insufficient motion representation due to compact 

In addition, limited by the capabilities of generative models and the internal GFVC mechanisms (e.g. 1) if the base picture is not updated, the texture reference for subsequent picture generation will be unchanged; and 2) compact representations maintain a fixed dimensionality regardless of scene variations, sometimes limiting their ability to capture complex motion patterns in dynamic environments.), the reconstructed face video quality sometimes may not be stable. Such instability in reconstruction quality mainly includes objectionable background distortions, and/or inaccurate local motions in the mouth and/or eyes, which may damage the feasibility of practical applications. To improve reconstruction robustness, the JVET GFVC AhG has investigated a series of image processing techniques for signal fusion and enhancement, including a scalable representation and layered reconstruction scheme~\cite{JVET-AH0110}, background segmentation/inpainting/fusion operations~\cite{JVET-AH0118,JVET-AI0194}, pupil position redirection~\cite{JVET-AI0137}, colour calibration~\cite{JVET-AL0156} and QP-adaptive feature enhancement~\cite{JVET-AL0102}.

\subsection{Implementations of GenerativeNN$\left (  \right )$ and TranslatorNN$\left (  \right )$}

As shown in Fig.~\ref{NN_functionalities}.(a), GenerativeNN$\left (  \right )$ specifies a neural network that may be used to generate output pictures using the fixed format of facial parameters and previously decoded output pictures. It should be emphasized that GenerativeNN$\left (  \right )$ does not employ a fixed network architecture. Instead, it refers to any generative model capable of synthesizing facial images with compact facial parameters. Taking CFTE/FOMM/DAC/FV2V as examples, GenerativeNN$\left (  \right )$ commonly consists of two components: a dense motion estimator and a generator. The dense motion estimator converts decoded facial parameters into dense motion fields, while the generator further combines the previously decoded output pictures with these transformed dense motion fields to produce the final synthesized face results.  

As shown in Fig.~\ref{NN_functionalities}.(b), TranslatorNN$\left (  \right )$ specifies a neural network that may be used to convert various formats of facial parameters signaled in the SEI message into a fixed format of parameters. As such, TranslatorNN$\left (  \right )$ could achieve the seamless translation between different facial representations and avoid the mismatch problem bewteen GFVC encoder and decoder. TranslatorNN$\left (  \right )$ is designed in a portable and light-weighted manner, which consists of a trans-encoder and a trans-decoder. In particular, each trans-en/decoder consists of 4 full-connected~(FC) layers with ReLU activation in between, and the dimension of FC layers and latent embedding is set to 256. 

\subsection{Encoding/Decoding Processes}
The SEI-based GFVC workflow is shown in Fig.~\ref{GFV_approach}, including the encoding and decoding processes. The SEI-based GFVC encoder performs feature analysis on the input face video to obtain compact facial representations via the analysis model, and then encapsulates these learned compact parameters according to the syntax defined by the GFV SEI message. Afterwards, these GFV SEI messages are packed together with the coded base pictures into VVC coded bitstreams. To further conform to the VVC Main Profile, the base pictures are coded in the $YC_{B}C_{R}$ 4:2:0 format.

The decoder receives the bitstream to reconstruct the face video. In particular, a VVC decoder with support for the GFV SEI message is utilized to decode the base pictures and parse the GFV SEI message. 
Since the GFVC synthesis is expected to proceed in 4:4:4 RGB format, the decoded base pictures should execute a color format conversion prior to the synthesis
Further, the parsed GFV SEI parameters and the converted base pictures are jointly fed into the synthesis model (i.e., deep generative network) to render the reconstructed video. 
It should be mentioned that a software implementation that streamlines all encoding and decoding processes has been proposed and made available by JVET GFVC AhG for further experimentation~\cite{JVET-AI0195}. Decoders that do not support the GFV SEI message would simply decode and display the result of the ordinary standardized core hybrid decoding process.

% \vspace{-1.1em}
\section{Experiments}

\begin{figure*}[tb]
\centering
\vspace{-1.2em}
\subfloat[User-specified Animation/Filtering]{\includegraphics[width=0.34\textwidth]{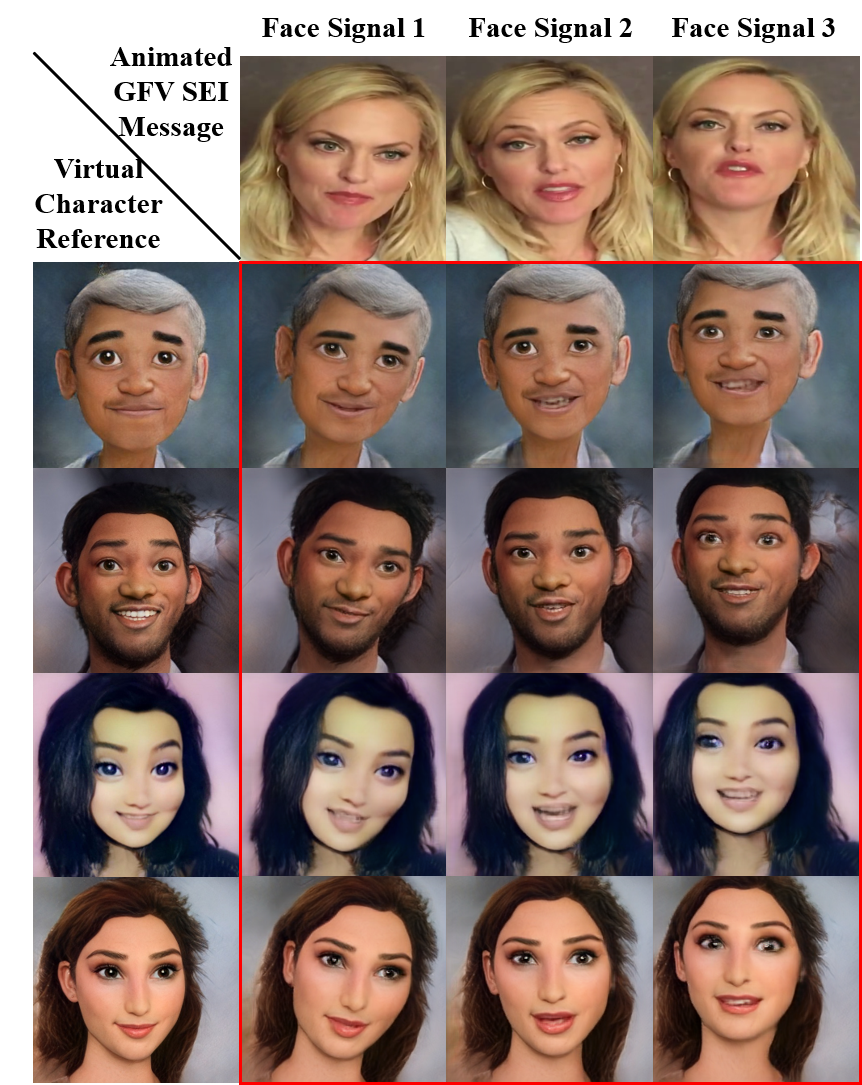}}
% \hspace{0.01\textwidth}
\hspace{1em}
\subfloat[Metaverse-related Interaction]{\includegraphics[width=0.46\textwidth]{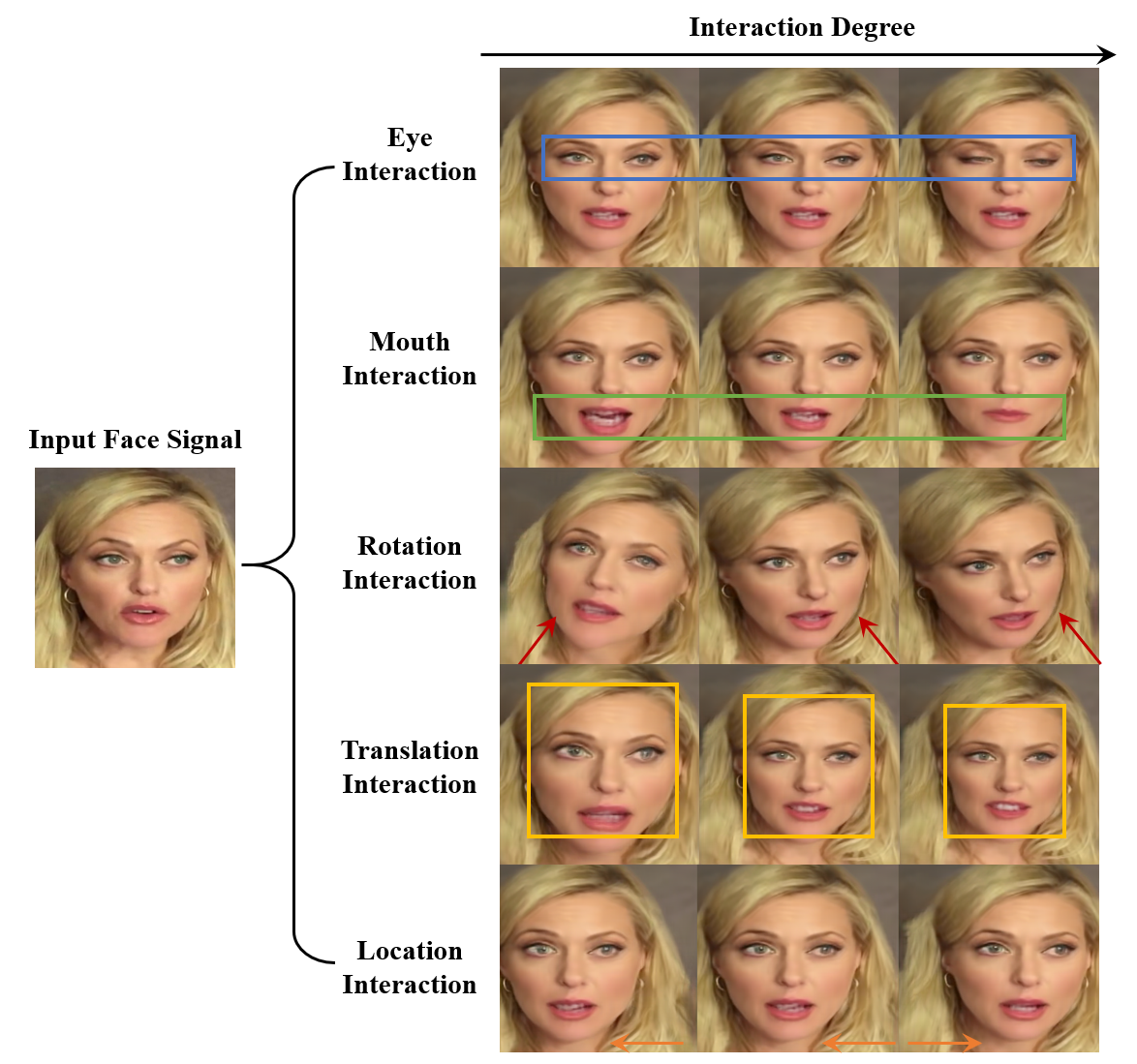}}
\caption{Illustration of SEI-based GFVC approach for diverse functionalities.} 
\label{SEI_functionalities} 
% \vspace{-2em}
\end{figure*}

\subsection{Experimental Settings}
\subsubsection{Implementation Specifics}
The GFV SEI message has been included in the working draft of VSEI version 4~\cite{JVET-AJ2006} and its implementation on top of the VVC reference software has been made available in the VTM SEI branch~\footnote{\href{https://vcgit.hhi.fraunhofer.de/jvet/VVCSoftware_VTM/-/merge_requests/2794}{VTM Software with GFV SEI Implementation}}. In addition, as the output of the JVET GFVC AhG activities, a GFVC software tool has been developed and made available on a GFVC git branch~\footnote{\href{https://vcgit.hhi.fraunhofer.de/jvet-ahg-gfvc/gfvc_v1}{JVET AhG16 GFVC Software Tools}}. Additionally, Python scripts are provided to link the GFVC software tool with the GFV SEI message implementation, such that the entire GFVC encoding/decoding processes using the GFV SEI approach can be executed in a streamlined manner. As for model training, each GenerativeNN$\left (  \right )$ (i.e., GFVC synthesis model) is trained on mixed data from VoxCeleb~\cite{Nagrani17} and CelebV~\cite{zhu2022celebvhq} training dataset for 100 epochs, whilst each TranslatorNN$\left (  \right )$ is trained on face features of VoxCeleb~\cite{Nagrani17} training sets for 400 epochs. It should be noted that these neural models and network interfaces have been adopted in the JVET GFVC AhG activities and available in the JVET GFVC software tools.

\subsubsection{Testing Conditions}
% Experiments of GFVC using a GFV SEI message implementation were conducted under the JVET GFVC AhG test conditions~\cite{JVET-AJ2035}, and a newly-added comparison anchor. 

Experiments of GFVC using a GFV SEI message implementation were compared to the VVC codec according to the GFVC AhG test conditions~\cite{JVET-AJ2035}, the adjusted VVC anchor and the latest learning-based video codec DCVC-FM~\cite{li2024neural}.
Four representative GFVC algorithms (i.e., CFTE~\cite{CHEN2022DCC}, FV2V~\cite{wang2021Nvidia}, DAC~\cite{ultralow} and FOMM~\cite{FOMM}) and a lightweight GFVC version CFTE-Lite are selected as testing algorithms. All testing sequences~\footnote{\href{https://vqa.lfb.rwth-aachen.de/index.php/apps/files/files/481617?dir=/jvet/ahg/gfvc}{JVET AhG16 Testing Sequences}} are available in two resolutions 256$\times$256 (Class A\&B) and 512$\times$512 (Class C\&D), where the total frame number of ClassA\&C and ClassB\&D sequences is 250 and 125, respectively. As shown in Fig.~\ref{CTC}, these test sequences span considerable diversity of face video content in terms of age, gender, ethnicity, appearance, expression, head position, head motion, camera motion, background, etc. They are pre-processed (i.e., cropped) from open-source high-quality high-resolution video, and therefore have smaller resolutions relative to the original video and have squared aspect ratio. 
The detailed experimental settings are listed as follows:
\begin{itemize}
\item{For the VVC anchor, VTM-22.2 reference software for VVC is used under low-delay B (LDB) configuration with quantization parameter (QP) set to 37, 42, 47 and 52.}
\item{For DCVC-FM, low delay configuration is used with quantization factor set to 30, 20, 10 and 0.}
\item{For the SEI-based GFVC tests, the first picture in each sequence is taken as the base picture and coded as an I picture using VTM-22.2 with QP values of 22, 32, 42 and 52. All other pictures are coded with the SEI-based GFVC approach, where the compact feature parameters are signaled via the GFV SEI message. It should be noted that in our experimentation, the coding frequency of the base picture is 1 time, i.e., the base picture is not additionally updated during the whole encoding process.}
\item{To obtain more aligned bitrate ranges and reference quality with the GFVC algorithms, the QP settings for the VVC anchor (LDB mode) have been further adjusted. By doing this, the bitrates of the adjusted VVC anchor are effectively reduced, creating a larger range overlap with the proposed SEI-based GFVC method. The corresponding QP settings for the two testing resolutions are listed: 1) For the 256$\times$256 resolution, the QP of an I picture is set at $\{22, 32, 42, 52\}$ and the QP of the subsequent frames are correspondingly fixed at $\{47, 47, 52, 57\}$. 2) For the 512$\times$512 resolution, the QP of an I picture is set at $\{22, 32, 42, 52\}$ and the QP of the subsequent frames are correspondingly fixed at $\{52, 52, 57, 57\}$.}

\end{itemize}

For objective quality evaluation of the reconstructed face video, we adopt Deep Image Structure and Texture Similarity (DISTS)~\cite{dists} and Learned Perceptual Image Patch Similarity (LPIPS)~\cite{lpips}, where a smaller score of these two measures indicates higher image quality.

\begin{table*}[t]
\vspace{-2em}
\centering
\caption{Model complexity and coding efficiency of different SEI-based GFVC approaches at the resolution of 256$\times$256 and 512$\times$512}  
\renewcommand\arraystretch{1.25}
\resizebox{0.75\linewidth}{!}{
\label{complexity}
\begin{tabular}{ccccccccc}
\hline
Resolution               & Side                     & Measure     & VVC              & CFTE    & FV2V    & DAC     & FOMM    & CFTE-Lite        \\ \hline
\multirow{6}{*}{256×256} & \multirow{3}{*}{Encoder} & Params (M)       & \textbackslash{} & 14.12         & 72.13         & 14.15        & 14.21         & 1.62               \\
                         &                          & kMACs/pixel      & \textbackslash{} & 15.10         & 120.28        & 16.20        & 19.49         & 1.84               \\
                         &                          & FPS              & 0.36             & 16.93         & 14.12         & 16.89        & 15.08         & 16.45              \\
                         & \multirow{3}{*}{Decoder} & Params (M)       & \textbackslash{} & 44.11         & 53.32         & 45.80        & 45.80         & 6.03               \\
                         &                          & kMACs/pixel      & \textbackslash{} & 1057.71       & 3211.12       & 1039.71      & 1039.71       & 172.40             \\
                         &                          & FPS              & 504.29           & 24.09         & 10.11         & 23.23        & 21.58         & 27.03              \\ \hline
\multirow{6}{*}{512×512} & \multirow{3}{*}{Encoder} & Params (M)       & \textbackslash{} & 14.12         & 72.13         & 14.15        & 14.21         & \textbackslash{}   \\
                         &                          & kMACs/pixel      & \textbackslash{} & 3.77          & 120.28        & 4.05         & 4.87          & \textbackslash{}   \\
                         &                          & FPS              & 0.11             & 9.08          & 8.07          & 9.07         & 8.56          & \textbackslash{}   \\
                         & \multirow{3}{*}{Decoder} & Params (M)       & \textbackslash{} & 44.11         & 53.32         & 45.80        & 45.80         & \textbackslash{}   \\
                         &                          & kMACs/pixel      & \textbackslash{} & 389.13        & 927.48        & 384.63       & 384.63        & \textbackslash{}   \\
                         &                          & FPS              & 156.52           & 15.39         & 7.80          & 14.51        & 14.34         & \textbackslash{}   \\ \hline
\end{tabular}
}
% \vspace{-2.5mm}
\end{table*}

\subsection{Performance Comparisons}

As illustrated in Fig.~\ref{RD}, we provide the rate-distortion performance comparisons among VVC, DCVC-FM and the proposed SEI-based GFVC apporach. Experimental results show that the proposed SEI-based GFVC approach can achieve promising rate-distortion performance in perceptual quality metrics like DISTS/LPIPS but perform poorly in pixel-level quality metrics like PSNR compared to the VVC codec according to the GFVC AhG test conditions, the adjusted VVC anchor and DCVC-FM. This result is consistent with the fact that the generative compression is designed in the feature domain rather than the pixel-level domain, so the reconstruction quality cannot achieve pixel-level alignment~\cite{10477607,chen2025generative}.

In terms of the rate-distortion performance of the original VVC anchor and the adjuested VVC anchor, it can be observed that at the resolution of 256$\times$256, the adjusted VVC anchor shows a small performance improvement. However, as the resolution increases to 512$\times$512, the performance advantage in the high bitrate range completely disappears, and the adjusted VVC anchor underperforms the original VVC anchor. As such, it can be concluded that the GFVC testing conditions from JVET-AJ2035~\cite{JVET-AJ2035} are relatively fair, and do not intentionally degrade the performance of the VVC codec to highlight the performance advantages of GFVC. Compared to both anchors and the DCVC-FM algorithm, our proposed SEI-based GFVC approach can achieve promising compression performance in higher bitrate ranges for the 256$\times$256 resolution. When the face video resolution is increased to 512$\times$512, the proposed SEI-based GFVC approach can achieve coding performance benefits for a wider range of bit rates. This is because the overhead of sending facial parameters via SEI messages becomes smaller relative to the VVC-coded base pictures as the video resolution increases. Such performance results confirm that SEI-based GFVC can provide promising coding performance for face video with resolutions at or above standard definition television resolution. Moreover, Fig.~\ref{RD} shows the compression efficiency of these SEI-based GFVC methods. In particular, DAC achieves superior compression efficiency compared to FV2V and CFTE, with FOMM demonstrating the least effective performance. This hierarchical ranking directly demonstrates that DAC produces the most compact 2D keypoints representation to achieve the best compression efficiency among these tested GFVC methods.

Fig.~\ref{subjective} provides subjective quality comparisons using VVC, DCVC-FM and the proposed SEI-based GFVC approaches at similar bit-rates. At ultra-low bitrates, VVC-reconstructed or DCVC-FM-reconstructed face images suffer from severe artifacts like blockiness, blurring, and loss of fine facial features due to aggressive compression. In contrast, the proposed SEI-based approaches can still reconstruct vivid faces with higher fidelity and provide much better texture details. This stark disparity underscores the efficacy of the proposed SEI-based GFVC approaches in mitigating quality degradation for face video at extremely low bitrate ranges. Moreover, it can be observed that SEI-based GFVC methods sometimes result in imprecise reconstruction of local details like mouth and eye movements. For example, FV2V exhibits deviations in reconstructing the mouth region, while CFTE shows inaccuracies in the eye area, both of which are less effective than DAC in achieving accurate local reconstruction.

\subsection{Functionalities Examples }
In addition to ultra-low bandwidth video communication, this SEI-based GFVC approach enables various purposes, such as user-specified animation/filtering and metaverse-related applications, with capabilities that cannot be directly supported by the core VVC codec. As illustrated in Fig.~\ref{SEI_functionalities}, the proposed approach gives the sender control over the quality of animated or filtered face pictures, and it can also support parameter modification in the GFV SEI message for expression or posture interactions in a virtual world. 

In particular, using a virtual anime face for texture reference, a series of GFV SEI messages indicating realistic facial motion and expression can animate this reference image to generate face frames. This technology offers virtual character-based video animation and filtering, safeguarding user privacy in contexts like virtual live-streaming and secure video chatting. Moreover, the facial parameters embedded within the GFV SEI message correspond distinctly to various facial features and motions, such as mouth movements, eye blinks, head rotations, translations, and locations. This facilitates semantic-level controls and friendly interactions over the reconstructed face video, opening up exciting prospects for future video conferencing and metaverse-related activities.

\begin{table*}[t]
% \vspace{-1.3em}
\centering
\caption{Average bit ratio of different SEI-based GFVC approaches with base picture QP =22, 32, 42, 52 at the resolution of 256$\times$256 and 512$\times$512}  
\renewcommand\arraystretch{1.15}
\resizebox{0.75\linewidth}{!}{
\label{bitratio}
\begin{tabular}{cccccc}
\hline
\multirow{2}{*}{\begin{tabular}[c]{@{}c@{}}SEI-based\\ GFVC Methods\end{tabular}} & \multirow{2}{*}{Resolution} & \multicolumn{4}{c}{GFVC Bit Ratio (Base Picture v.s. All Pictures)}                                                                                                                                                                                        \\ \cline{3-6} 
                                                                                  &                             & \begin{tabular}[c]{@{}c@{}}Base Picture \\ QP=22\end{tabular} & \begin{tabular}[c]{@{}c@{}}Base Picture\\ QP=32\end{tabular} & \begin{tabular}[c]{@{}c@{}}Base Picture\\ QP=42\end{tabular} & \begin{tabular}[c]{@{}c@{}}Base Picture\\ QP=52\end{tabular} \\ \hline
\multirow{2}{*}{CFTE}                                                             & 256×256                     & 51\%                                                          & 30\%                                                         & 15\%                                                         & 8\%                                                          \\
                                                                                  & 512×512                     & 72\%                                                          & 48\%                                                         & 25\%                                                         & 12\%                                                         \\
\multirow{2}{*}{FV2V}                                                             & 256×256                     & 53\%                                                          & 31\%                                                         & 16\%                                                         & 8\%                                                          \\
                                                                                  & 512×512                     & 73\%                                                          & 49\%                                                         & 25\%                                                         & 12\%                                                         \\
\multirow{2}{*}{DAC}                                                              & 256×256                     & 58\%                                                          & 36\%                                                         & 19\%                                                         & 10\%                                                         \\
                                                                                  & 512×512                     & 78\%                                                          & 54\%                                                         & 30\%                                                         & 15\%                                                         \\
\multirow{2}{*}{FOMM}                                                             & 256×256                     & 68\%                                                          & 45\%                                                         & 24\%                                                         & 12\%                                                         \\
                                                                                  & 512×512                     & 73\%                                                          & 49\%                                                         & 27\%                                                         & 13\%                                                         \\
CFTE-Lite                                                                         & 256×256                     & 53\%                                                          & 30\%                                                         & 15\%                                                         & 8\%                                                          \\ \hline
\end{tabular}
}
% \vspace{-3.5mm}
\end{table*}

\subsection{Complexity Analysis}

To show the model complexity and coding efficiency of SEI-based GFVC approaches, we compare the number of model parameters (Params/M), number of operations measured by Multiply–Accumulate (kMACs/pixel) as well as actual inference efficiency (FPS). Herein, we employ four original GFVC models (i.e., CFTE/FOMM/DAC/FV2V) and a lightweight GFVC model (i.e., CFTE-Lite) in this test. 
The overall encoding/decoding processes are both executed on Tesla V100-SXM2-32GB with 24 core CPUs (Intel(R) Xeon(R) Platinum 8163 CPU@2.50GHz). In addition, the testing sequence has 250 frames at the two different resolutions 256$\times$256 and 512$\times$512. To avoid the bias, the actual inference time at the encoder and decoder sides are both averaged with different QP settings according to JVET GFVC AhG test conditions~\cite{JVET-AJ2035}. 

As shown in Table \ref{complexity}, compared with the traditional VVC codec, SEI-based GFVC schemes can significantly reduce the actual inference time of encoder side, while the computational complexity and inference efficiency are both increased for decoding. As for these GFVC models, their decoders are usually faced with more parameters and higher complexity than their encoders, since these decoders are designed with sophisticated architecture for accurate motion estimation and vivid signal generation. In addition, with the increase of video resolution, the value of kMACs-per-pixel can be reduced. Moreover, the lightweight GFVC version CFTE-Lite can achieve good trade-offs among model size, complexity and actual inference time whilst maintaining promising RD performance. In particular, CFTE-Lite can greatly reduce the computational complexity of CFTE encoder (from 15.10 kMACs/pixel to 1.84 kMACs/pixel) and decoder (from 1057.71 kMACs/pixel to 172.40 kMACs/pixel) at the video resolution of 256$\times$256. 

\begin{figure}[t]
\centering
% \vspace{-2.8em}
\subfloat[CFTEtoDAC]{\includegraphics[width=0.16\textwidth]{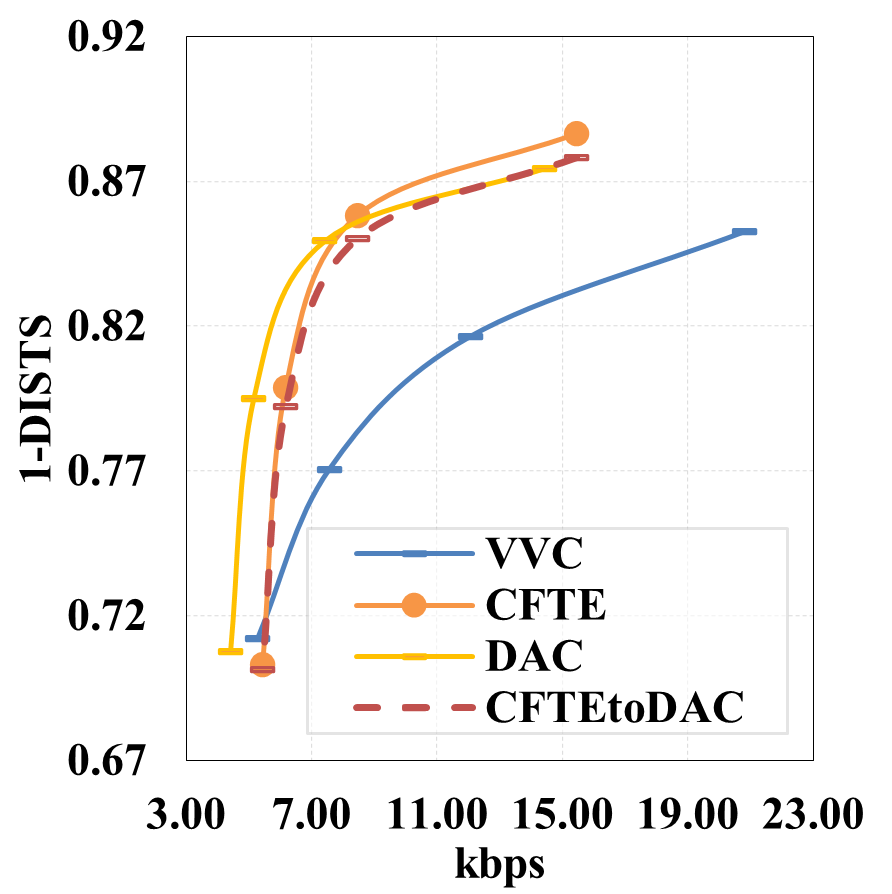}}
% \hspace{3em}
\subfloat[CFTEtoFOMM]{\includegraphics[width=0.16\textwidth]{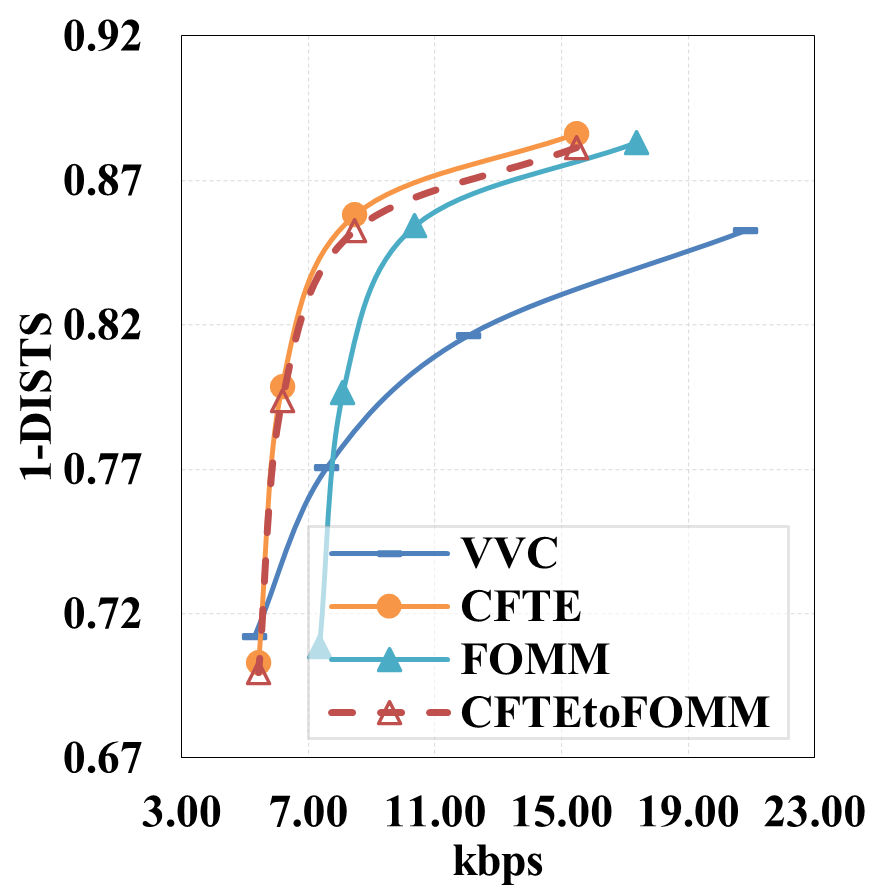}}
\subfloat[CFTEtoFV2V]{\includegraphics[width=0.16\textwidth]{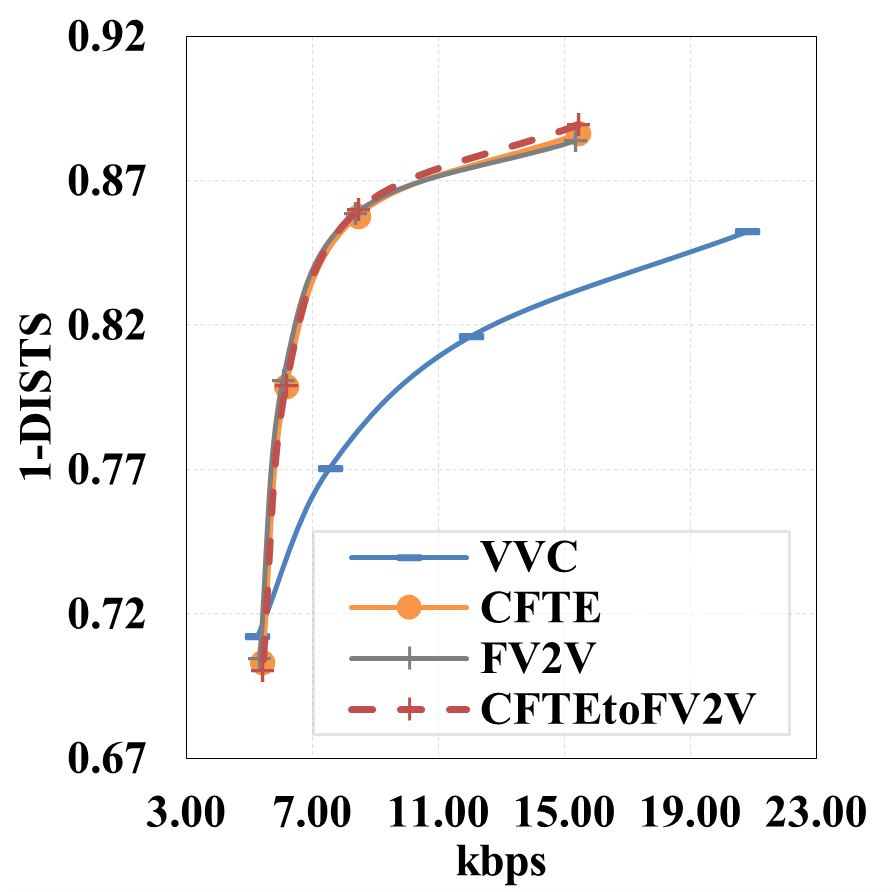}}
\caption{Average Rate-DIDTS performance of using TranslarorNN$\left (  \right ) $ for CFTEtoDAC, CFTEtoFOMM and CFTEtoFV2V scenarios at 256$\times$256 and 512$\times$512 resolutions.}   % 
\label{translatorNN}
% \vspace{-1.2em}
\end{figure}

\begin{figure}[tb]
% \vspace{-1em}
\centering
\includegraphics[width=0.43 \textwidth]{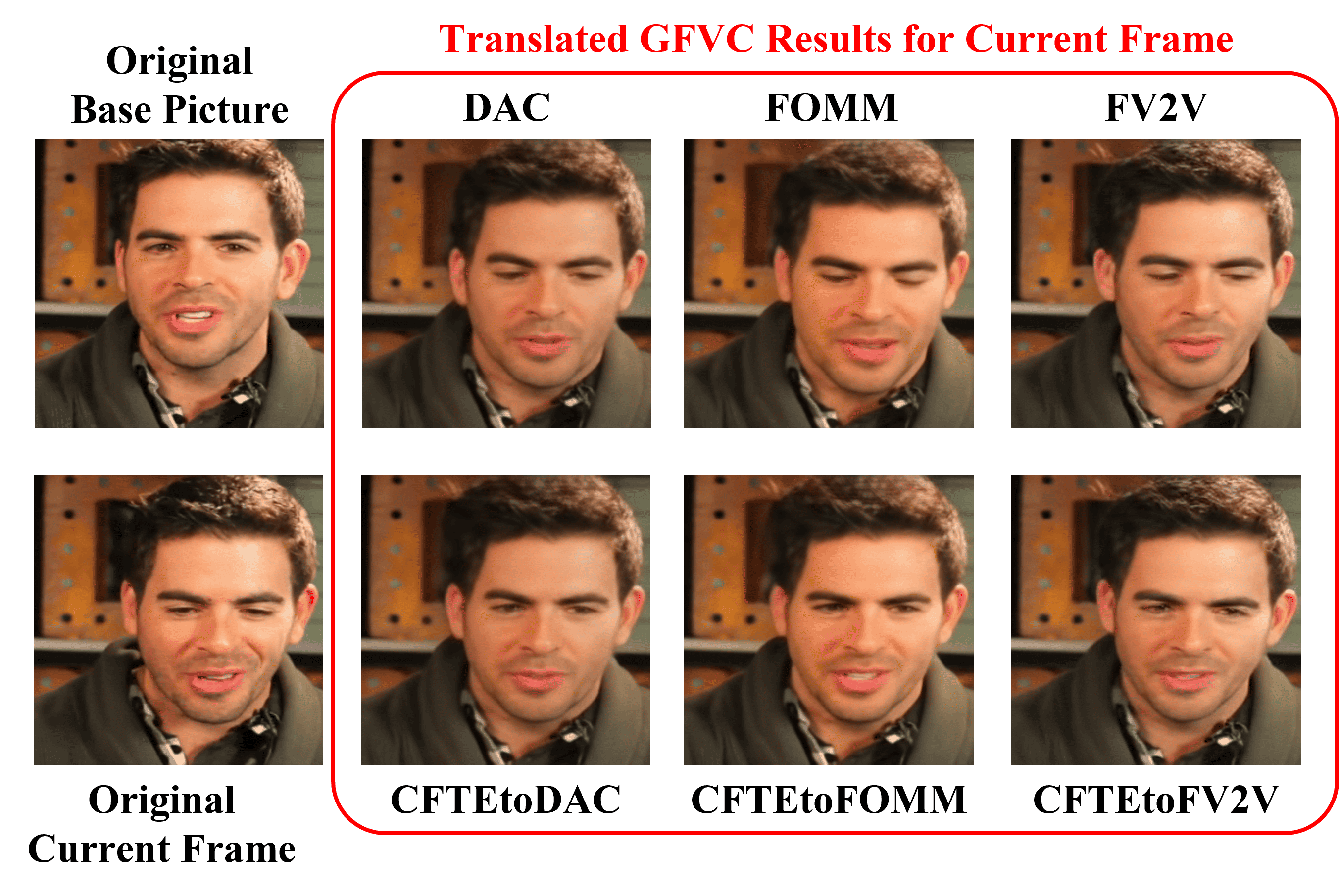}
\caption{Subjective quality comparisons of translated GFVC systems using TranslatorNN$\left (  \right ) $. }
% \vspace{-1em}
\label{translated_resuls} 
\end{figure}

\subsection{TranslatorNN$\left (  \right ) $  Interoperability Analysis}
To evaluate the effectiveness of TranslatorNN$\left (  \right ) $, we measure the RD performance of translated GFVC systems, where the encoder and decoder are equipped with different GFVC models utilizing different feature representations. The detail discussion and analysis of TranslatorNN can be referred to~\cite{JVET-AG0048,JVET-AL0147,yin2024parametertranslator}. Herein, we take the CFTE algorithm as a test model to validate the feasibility of TranslatorNN$\left (  \right ) $ in `CFTEtoAny' scenario. Specifically, CFTE~\cite{CHEN2022DCC} is used as the encoder and DAC~\cite{ultralow}/FOMM~\cite{FOMM}/FV2V~\cite{wang2021Nvidia} are used as decoder, respectively. As illustrated Fig.~\ref{translatorNN}, these translated GFVC systems like ``CFTEtoDAC'', ``CFTEtoFOMM'' and ``CFTEtoFV2V'' show promising rate-DISTS performance and outperform VVC in a large margin. In other words, TranslatorNN$\left (  \right ) $ can enable interoperability between different types of GFVC algorithms and avoid mismatch between the GFVC encoder and decoder. Moreover, TranslatorNN$\left (  \right ) $ provides possibility to further improve RD performance. For example, for FOMM decoding system, the RD performances can be largely improved when a CFTE encoding system is utilized for less bit-rate consumption. The subjective quality of translated GFVC results and their original results are shown in Fig.~\ref{translated_resuls}. It can be observed that, with the aid of TranslatorNN$\left (  \right ) $,  GFVC decoder can achieve vivid generations despite GFVC encoder mismatches.

\subsection{Base-picture Coding Frequency Analysis}

In the GFVC framework, the decision to update the base picture can be modeled as an optimization problem. Maintaining a fixed base picture achieves minimal bitrate consumption, but may degrade texture reference and reconstruction quality over time due to temporal variations. Conversely, periodic base picture updates incur higher coding costs but improve reconstruction fidelity.
%It should be mentioned that in our above experimentation, the coding frequency of the base picture is 1 time, i.e., the base picture is not additionally updated during the whole encoding process. This is because the employed test sequences (125 or 250 frames) are relatively short, and and the changes in temporal motion between frames are minimal. Therefore, frequent updates of base pictures are not suitable in such a certain circumstance.

\begin{figure}[tb]
\centering
% \vspace{-2.8em}
\subfloat[256$\times$256: Rate-DISTS]{\includegraphics[width=0.24\textwidth]{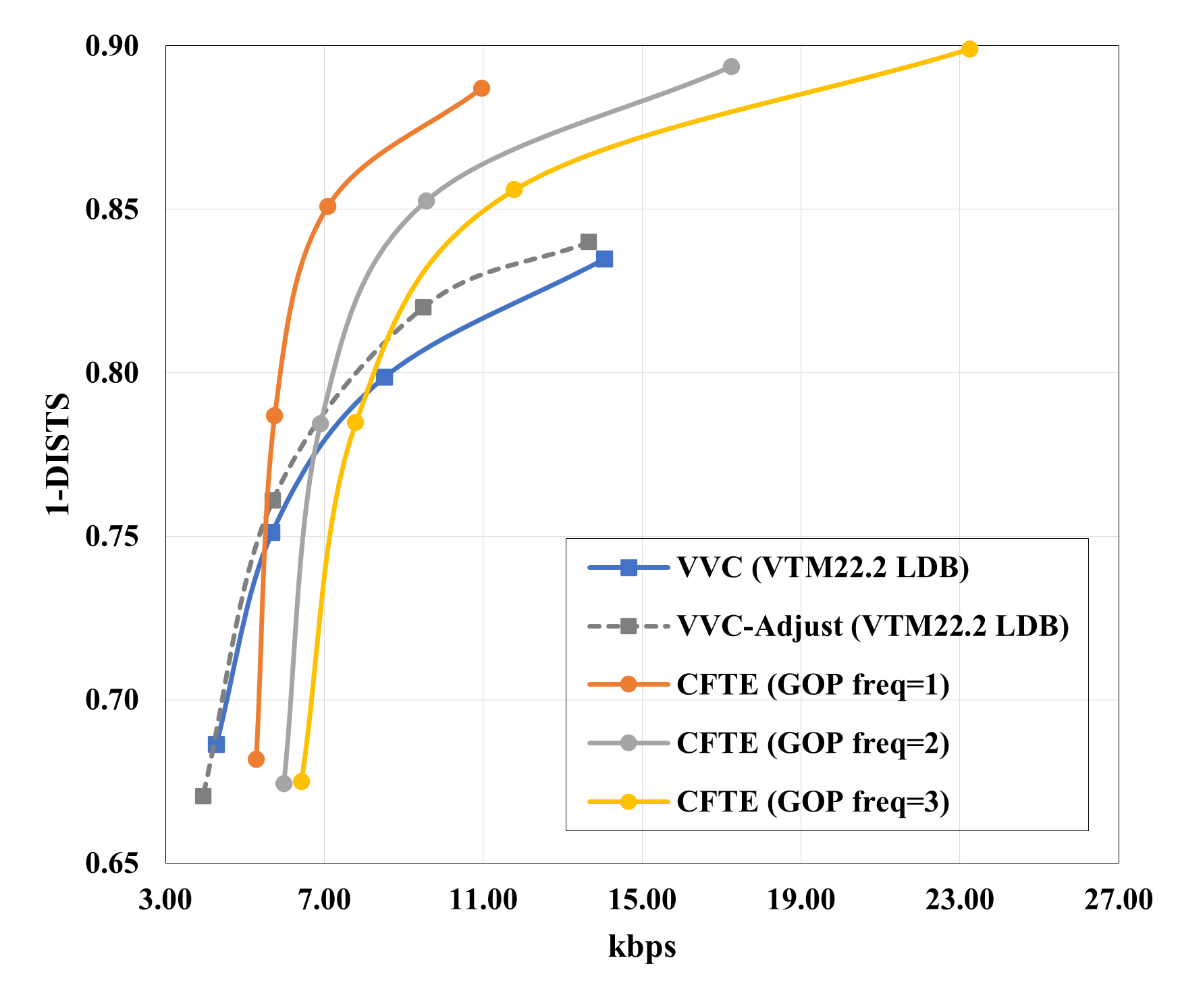}}
\hspace{0.05mm}
\subfloat[512$\times$512: Rate-DISTS]{\includegraphics[width=0.24\textwidth]{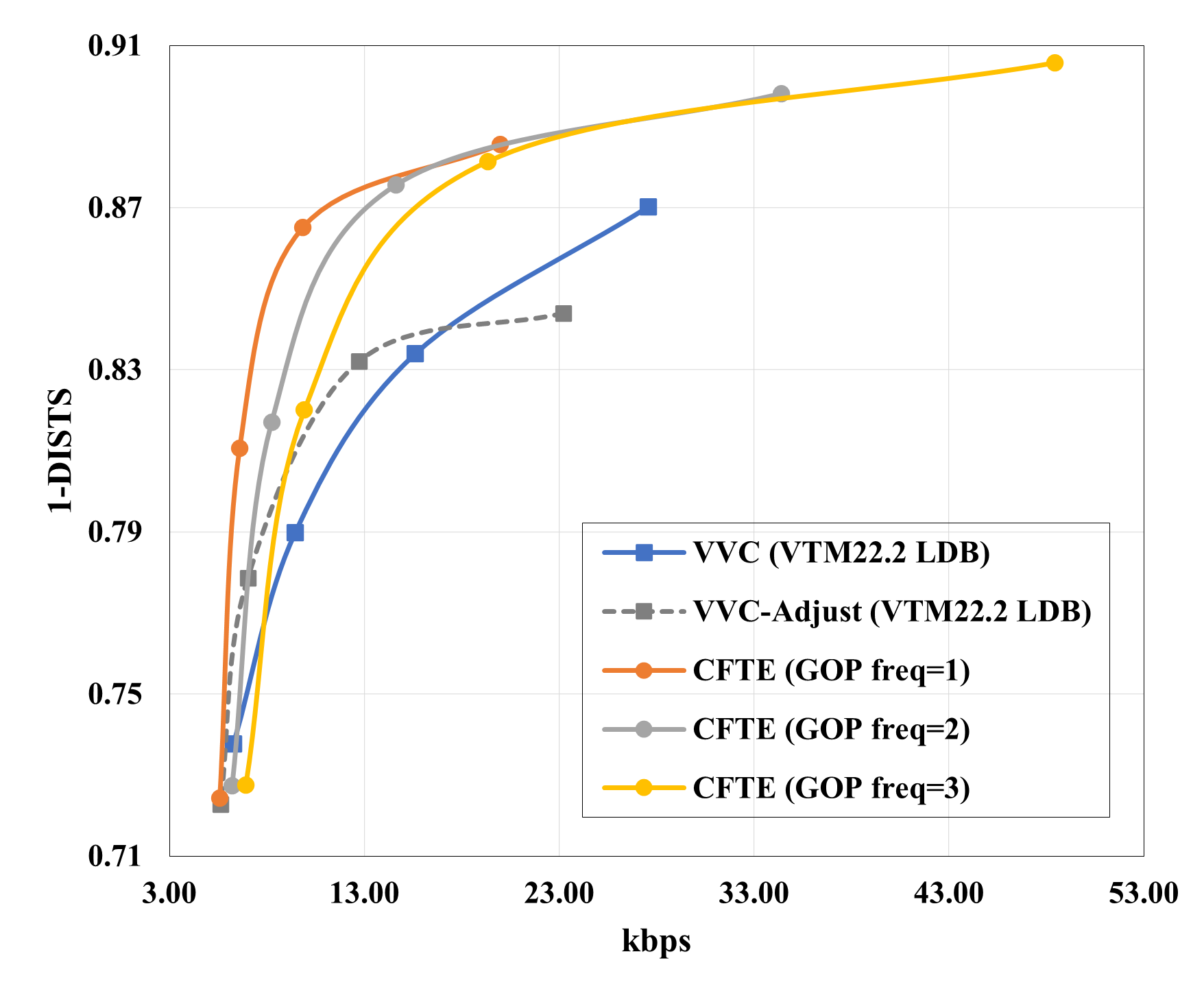}}
\caption{RD performance comparisons of VVC/VVC-adjust and the SEI-based CFTE method using different base-picture coding frequencies for both 256$\times$256 and 512$\times$512 resolutions.
} % 
\label{gop-rd}
% \vspace{-1.2em}
\end{figure}

To better discuss this point, we have added additional experimental analysis regarding base-picture coding frequency. Specifically, we use the CFTE algorithm as a test model and increase the base-picture encoding frequency from one to two and three times, and finally compare their RD results. As shown in Fig.~\ref{gop-rd}, when the base-picture encoding frequency increases, the reconstruction of subsequent inter frames could be provided with better texture references, which can improve the objective quality of the reconstructed frames. However, the bit cost of increasing the base picture encoding frequency is huge, and it is difficult to achieve a good rate-distortion trade-off. Moreover, as reported in~\cite{ultralow,icip2022zhao,chen2023csvt}, compared with directly dividing the video into different GOPs to obtain the base picture, dynamically deciding which video frame is the base picture based on the temporal motion changes may be more intelligent and can facilitate the RD performance optimization. Thus, there is room for improvement in exploring the base picture update mechanism within the GFVC framework as a direction for future work.

\begin{figure}[t]
\centering
\vspace{-2.8em}
\subfloat[256$\times$256 Resolution]{\includegraphics[width=0.245\textwidth]{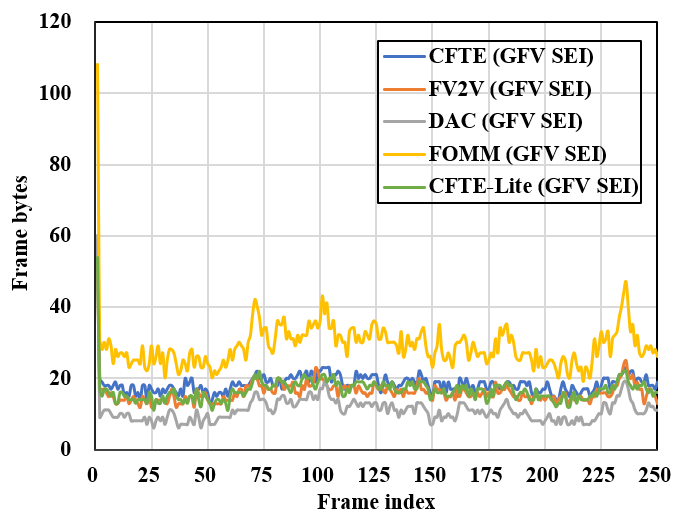}}
% \hspace{3em}
\subfloat[512$\times$512 Resolution]{\includegraphics[width=0.245\textwidth,]{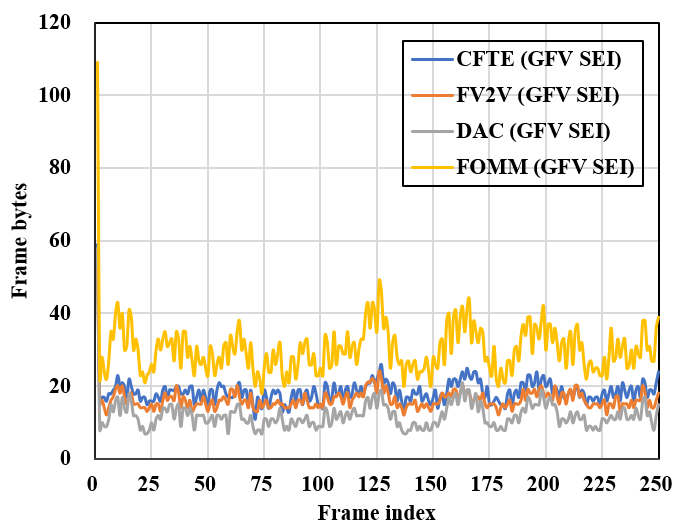}}
\caption{Illustration of bit fluctuation at picture level for  SEI-based GFVC approaches at 256$\times$256 and 512$\times$512 resolutions.} % 
\label{bit_fluctuation}
% \vspace{-1.2em}
\end{figure}

\subsection{Latency Analysis}
Herein, we have provided latency evaluation results and analyzed the latency of different SEI-based GFVC approaches from the following aspects: average bit ratio (base picture v.s. all pictures) and bit fluctuation at picture level.
 
Table \ref{bitratio} provides the average bit ratio that is used for the base picture and all pictures at different base picture QP settings. It can be revealed that the bit consumption of SEI-based GFVC approaches is more susceptible to the QP and resolution of base picture instead of GFV SEI messages. Specifically, as the QP of the base picture decreases or the resolution of the base picture increases, the bit ratio of base picture v.s. all pictures will both gradually increase. In addition, Fig.~\ref{bit_fluctuation} shows picture-level bit fluctuation for SEI-based GFVC approaches for two test sequences with 250 frames but different resolutions. It should be noted that the bit consumption of base picture is not counted but the bit consumption of its SEI messages is counted. As illustrated in Fig.~\ref{bit_fluctuation}, the proposed SEI-based GFVC approaches can stably allocate coding bits to all pictures without any obvious bit fluctuation. 

It should be mentioned that the proposed SEI-based GFVC approach follows a traditional low-delay setting in which the coding order and the display order are the same, and thus can well ensure high efficiency and low latency in video communication because SEI usually refers to additional data or features aimed at enriching the primary content, and these pieces of information are commonly transmitted and processed with high efficiency and low latency. As for the possible latency introduced by base-picture bitrate fluctuation, well-known techniques such as efficient buffer management, adaptive bitrate algorithms and network optimization can be employed. Such practices have long been successfully deployed in video conferencing applications to reduce the impact of bitrate fluctuations on latency in real-time video applications and are equally applicable to GFVC.

% \vspace{-1em}
\section{Conclusions and Discussions}

Generative AI has dramatically changed the acquisition of digital content and driven forward the innovative progress of visual data compression. However, these promising generative coding technologies  face the challenge of providing a standardized way to represent various feature formats, thus greatly limiting their practical deployment and breadth of applications. To address this, this paper discusses how to define a unified high-level syntax structure for different GFVC representations and insert it into a VVC bitstream as SEI messages, and can incorporate different GFVC algorithms in a standardized workflow. The simulation results demonstrate the effectiveness of the proposed SEI-based GFVC approach in delivering promising performance and diverse functionalities. 
The proposed GFV and GFVE SEI messages will be included in a new version of the VSEI standard that is expected to be finalized in the near future, and new versions of major video coding standards including VVC, HEVC and AVC will provide interface to enable the inclusion of GFV/GFVE SEI messages into VVC-, HEVC-, and AVC-coded bitstreams. This reflects the successful collaboration effort of multiple companies and research institutes to standardize generative video coding techniques within the JVET.

As for potential future work related to GFVC technique and its standardization, more robust model and stable reconstruction quality may improve the feasibility of practical applications. Besides, it is necessary to further reduce model complexity especially for the GFVC decoder and reduce deployment cost especially on mobile devices. 
The possible latency/delay problems in the real-time system also need to be solved to ensure the smoothness and stability of the playback.
Another promising direction is to systematically investigate the quality evaluation of generative coding, since the existing pixel-level or perceptual-level quality measures are not entirely suitable to reflect the quality of reconstructed video. Furthermore, it is envisioned that the proposed approach could inspire the development of future generative video coding technologies beyond face video. In particular, it is straightforward to extend the proposed standardization scheme from talking face video to human moving body video or even some natural-scene video that possess strong prior knowledge in structure of motion and texture, which can facilitate the generative video coding towards diverse scenarios and collaborated communication.

\bibliographystyle{IEEEtran}
\bibliography{bare_jrnl}

\begin{IEEEbiography}
[{\includegraphics[width=1in,height=1.25in,clip,keepaspectratio]{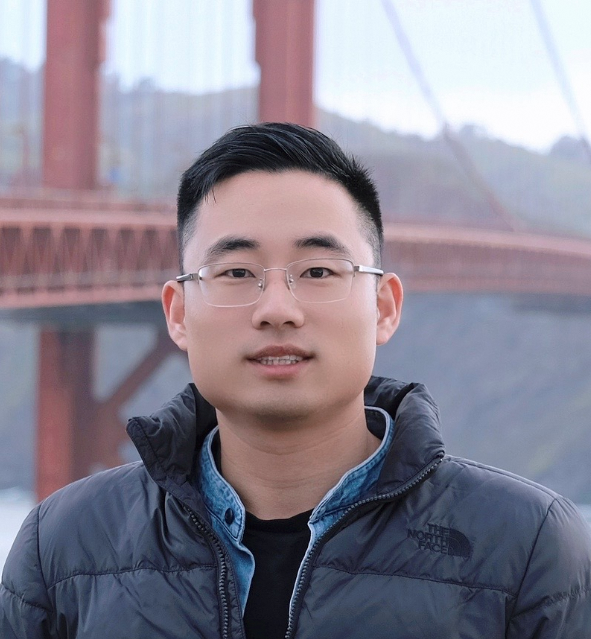}}]{Bolin Chen} (Member, IEEE) received the B.S. degree in communication engineering from Fuzhou University in July 2020 and the Ph.D. degree in computer science from City University of Hong Kong in February 2025, respectively. He is currently a senior algorithm engineer at Alibaba DAMO Academy, and is also a joint postdoctoral fellow at Alibaba DAMO Academy and Fudan University. Prior to this, he was a research assistant and a short-term postdoctoral fellow at City University of Hong Kong. His research interests include video compression, quality assessment and multimedia processing.
\end{IEEEbiography}

\vspace{-3em}
\begin{IEEEbiography}[{\includegraphics[width=1in,height=1.25in,clip,keepaspectratio]{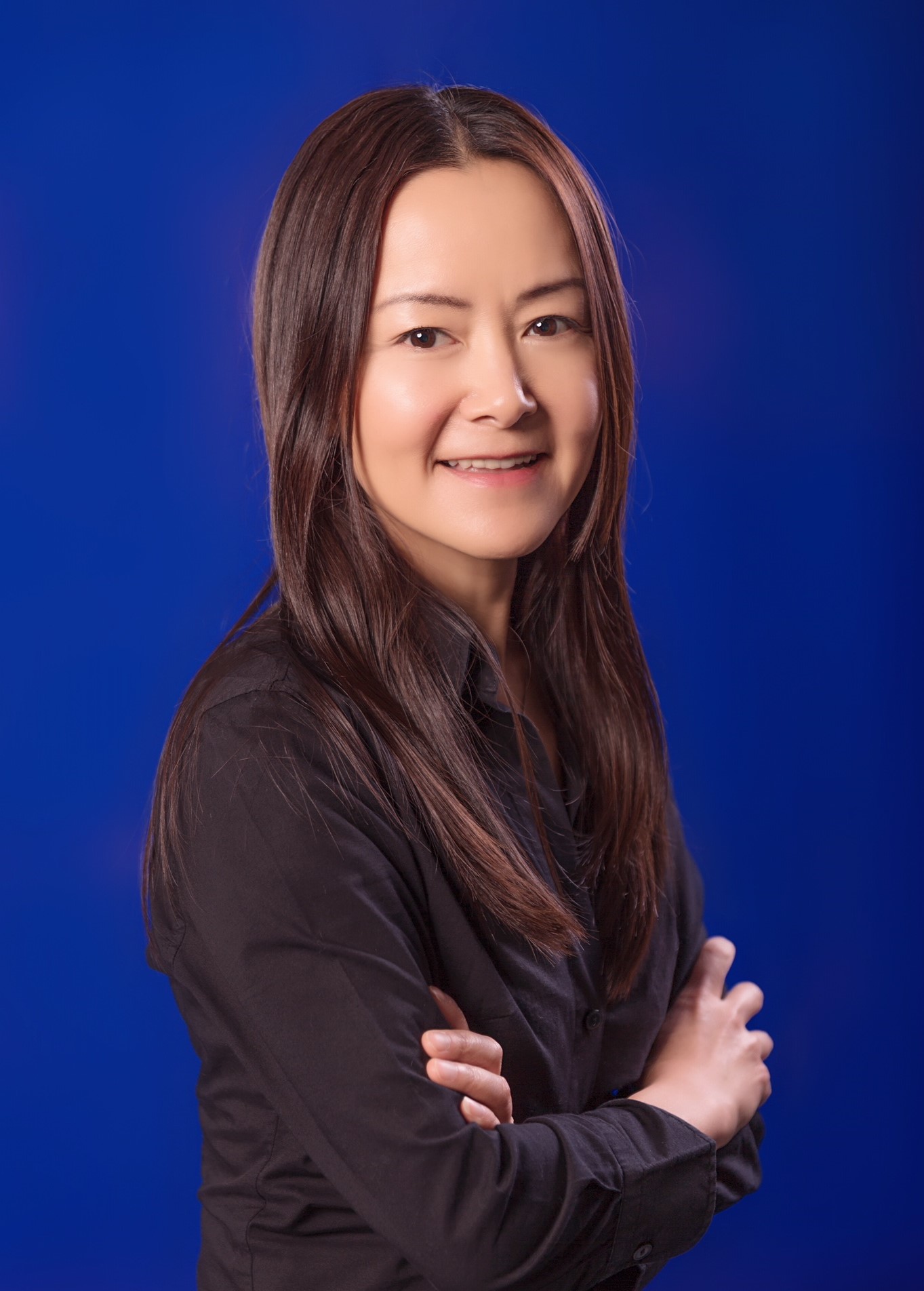}}]{Yan Ye} (Senior Member, IEEE) received the B.S. and M.S. degrees in electrical engineering from the University of Science and Technology of China in 1994 and 1997, respectively, and the Ph.D. degree in electrical engineering from the University of California at San Diego, in 2002. She is currently the head of Video Technology Lab at Alibaba Damo Academy, Alibaba Group U.S., Sunnyvale, CA, USA, where she oversees multimedia standards development, video codec implementation, and AI-based video research. Prior to Alibaba, she was with the Research and Development Labs, InterDigital Communications, Image Technology Research, Dolby Laboratories, and Multimedia Research and Development and Standards, Qualcomm Technologies, Inc. She has been involved in the development of various video coding and streaming standards, including H.266/VVC, H.265/HEVC, scalable extension of H.264/MPEG-4 AVC, MPEG DASH, and MPEG OMAF. She has published more than 80 papers in peer-reviewed journals and conferences. Her research interests include advanced video coding, processing and streaming algorithms, real-time and immersive video communications, AR/VR, and deep learning-based video coding, processing, and quality assessment. 
\end{IEEEbiography}

\vspace{-2em}
\begin{IEEEbiography}[{\includegraphics[width=1in,height=1.25in,clip,keepaspectratio]{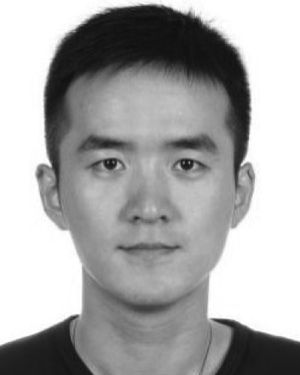}}]{Jie Chen} received B.S. and M.S. degrees in information and communication engineering form Zhejiang University, Hangzhou, China, in 2009 and 2012, respectively. He is currently with Video Technology Lab, Damo Academy, Alibaba Group, Beijing, China. Before joining Alibaba in 2019, he was with Beijing Samsung Telecom R\&D Center, Samsung Electronics from 2012 to 2018. He has been actively involved in the development of various video coding standards, including H.274/VSEI, H.266/VVC, AVS3 and AVS2 standard. His research interests include image and video coding and processing.
\end{IEEEbiography}

\vspace{-1cm}
\begin{IEEEbiography}[{\includegraphics[width=1in,height=1.25in,clip,keepaspectratio]{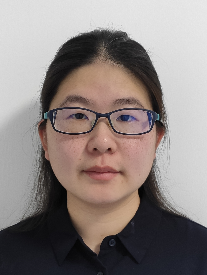}}]{Ru-Ling Liao} received the B.S. degree and M.S. degree in Computer Science and Engineer from National Chao Tung University, Hsinchu, Taiwan, in 2013 and 2016, respectively. She has been participated in the development of VVC standard since 2016. After 2019, she joined Alibaba Inc., Beijing, China. She is currently a senior engineer, and her research interests include video processing and neural network-based video compression technology.
\end{IEEEbiography}

\vspace{-4em}
\begin{IEEEbiography}[{\includegraphics[width=1in,height=1.25in,clip,keepaspectratio]{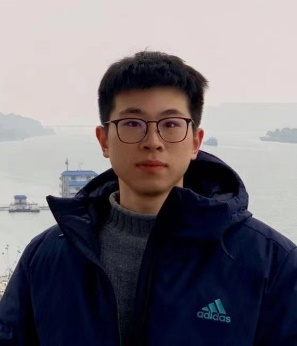}}]{Shanzhi Yin}  received the B.E. degree in communication engineering from Wuhan University of Technology, Wuhan, China, in 2020, and the M.S.
degree in information and communication engineering from Harbin Institute of Technology, Shenzhen, China, in 2023. He is currently pursuing the Ph.D. degree with the Department of Computer Science,
City University of Hong Kong. His research interests include video compression and generation.
\end{IEEEbiography}

\vspace{-4em}
\begin{IEEEbiography}[{\includegraphics[width=1in,height=1.25in,clip,keepaspectratio]{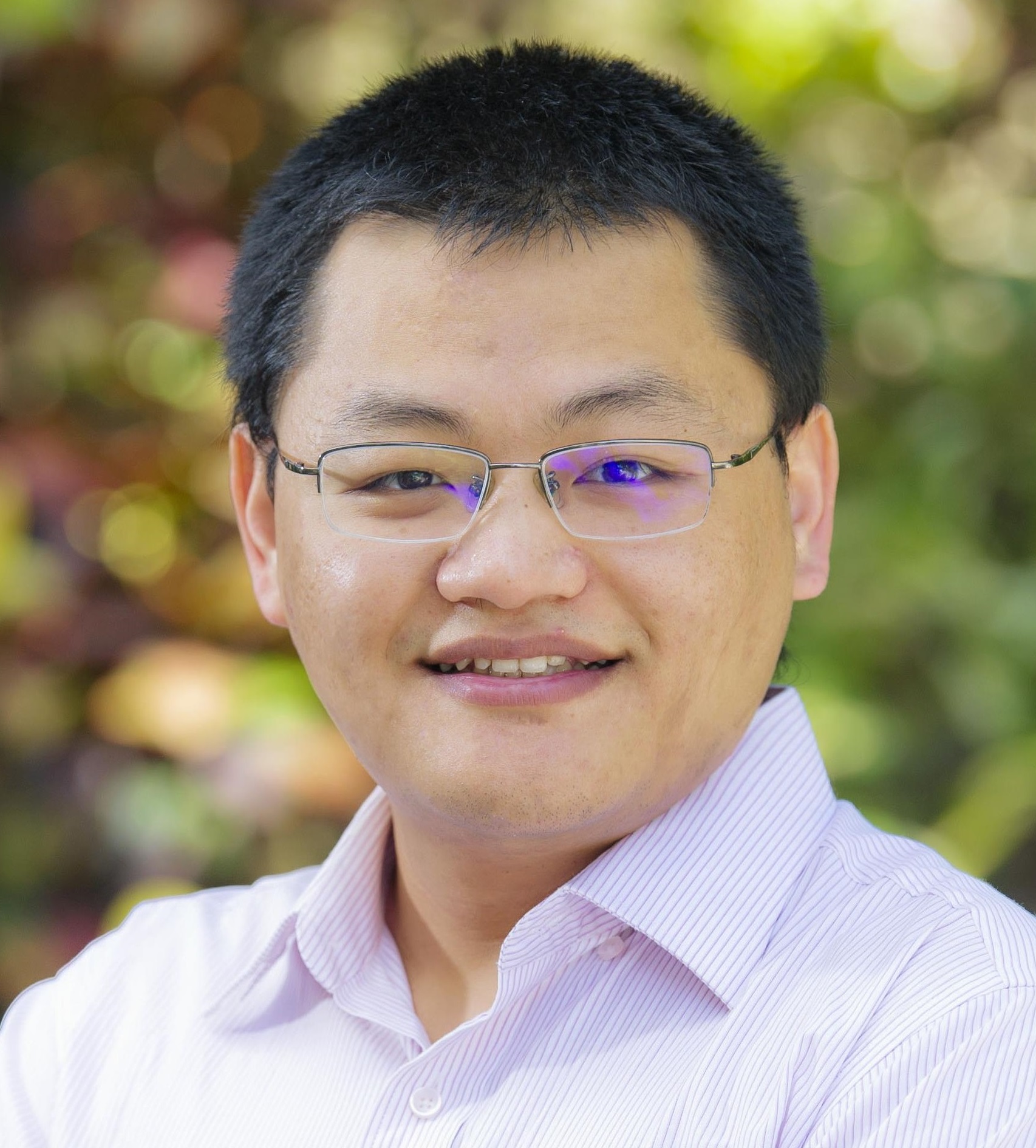}}]{Shiqi Wang} (Senior Member, IEEE) received the B.S. degree in computer science from the Harbin Institute of Technology in 2008 and the Ph.D. degree in computer application technology from Peking University in 2014. From 2014 to 2016, he was a Post-Doctoral Fellow with the Department of Electrical and Computer Engineering, University of Waterloo, Waterloo, ON, Canada. From 2016 to 2017, he was a Research Fellow with the Rapid-Rich Object Search Laboratory, Nanyang Technological University, Singapore. He is currently a Professor with the Department of Computer Science, City University of Hong Kong. He has proposed more than 50 technical proposals to ISO/MPEG, ITU-T, and AVS standards, and authored or coauthored more than 300 refereed journal articles/conference papers. His research interests include video compression, image/video quality assessment, and image/video search and analysis. He received the Best Paper Award from IEEE VCIP 2019, ICME 2019, IEEE Multimedia 2018, and PCM 2017. His coauthored article received the Best Student Paper Award in the IEEE ICIP 2018. He was a recipient of the 2021 IEEE Multimedia Rising Star Award in ICME 2021. He serves as an Associate Editor for \textsc{IEEE Transactions on Circuits and Systems for Video Technology}, \textsc{IEEE Transactions on Image Processing}, \textsc{IEEE Transactions on Multimedia} and \textsc{IEEE Transactions on Cybernetics}. 
\end{IEEEbiography}

% \vspace{-4em}
\begin{IEEEbiography}
[{\includegraphics[width=1in,height=1.25in,clip,keepaspectratio]{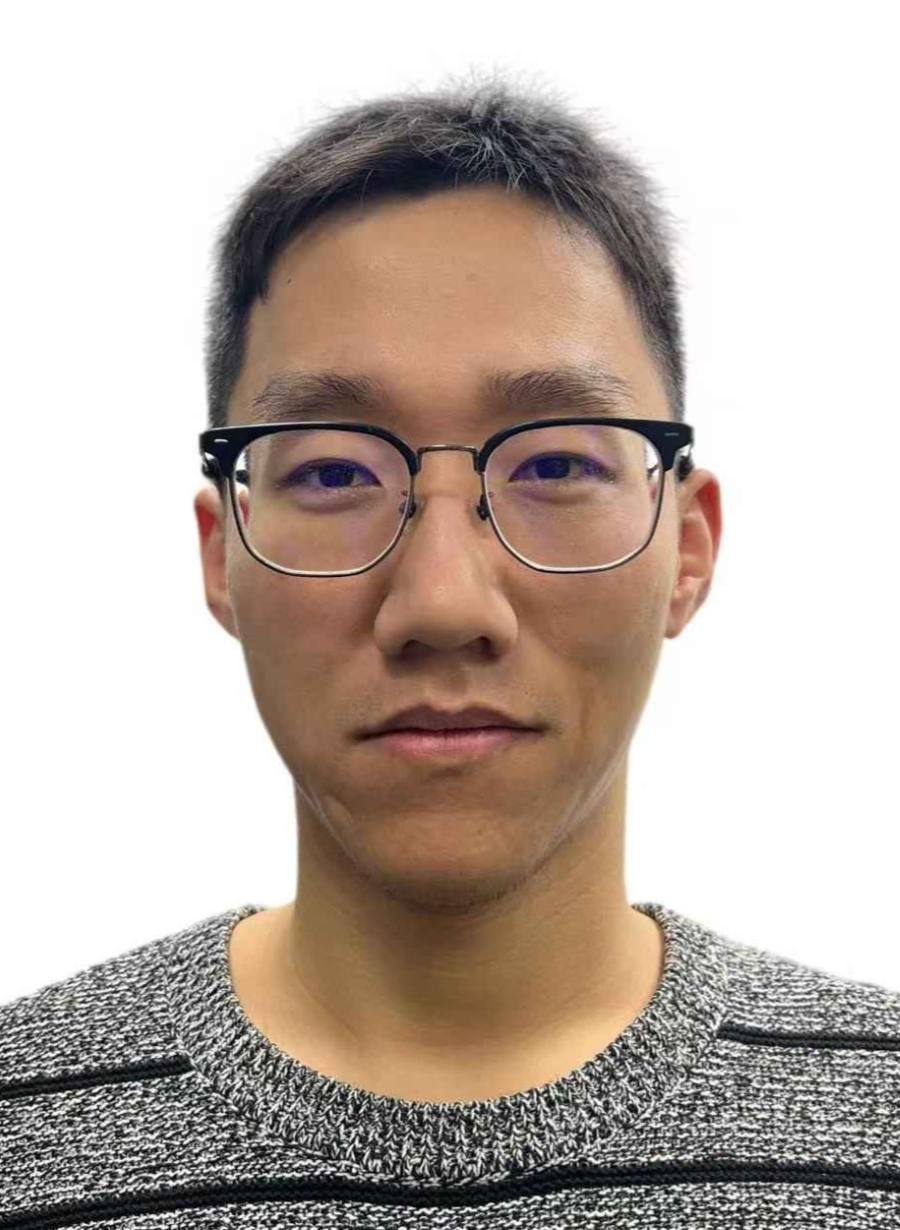}}]{Kaifa Yang}
received the B.S. degree in electronic and information engineering from Xi'an Jiaotong University, xi'an, China, in 2021. He is currently working toward the Ph.D degree in information and communication engineering at Shanghai Jiao Tong University, Shanghai, China. His research interests include visual quality assessment and quality-related optimization.
\end{IEEEbiography}

% \vspace{-4em}
\begin{IEEEbiography}
[{\includegraphics[width=1in,height=1.25in,clip,keepaspectratio]{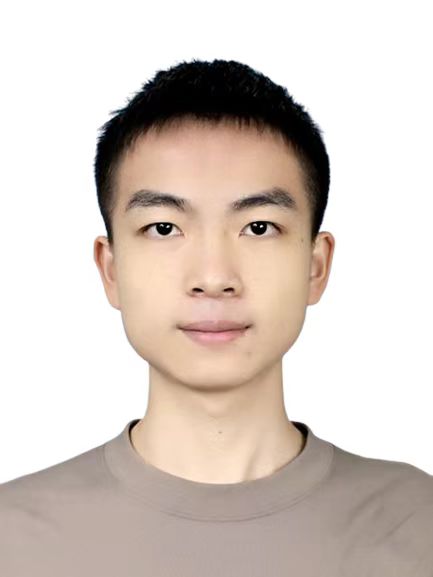}}]{Yue Li} received the B.S. degree in electronic information science and technology from Nankai University, Tianjin, China, in 2022. He is currently pursuing the Ph.D. degree in information and communication engineering at Shanghai Jiao Tong University, Shanghai, China. His research interests include multimedia transmission and forward error correction.
\end{IEEEbiography}

% \vspace{-4em}
\begin{IEEEbiography}
[{\includegraphics[width=1in,height=1.25in,clip,keepaspectratio]{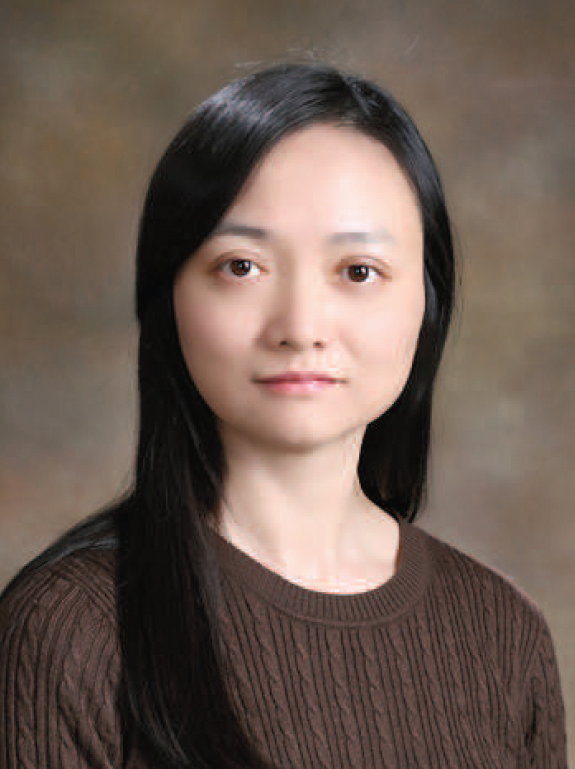}}]{Yiling Xu}
received the BS, MS, and PhD degrees from the University of Electronic Science and Technology of China, in 1999, 2001, and 2004 respectively. From 2004 to 2013, she was a senior engineer with the Multimedia Communication Research Institute, Samsung Electronics Inc., South Korea. She joined Shanghai Jiao Tong University, where she is currently a professor in the areas of multimedia communication, 3D point cloud compression and assessment, system design, and network optimization. She is the associate editor of the IEEE Transactions on Broadcasting. She is also an active member in standard organizations, including MPEG, 3GPP, and AVS.
\end{IEEEbiography}

% \vspace{-4em}
\begin{IEEEbiography}
[{\includegraphics[width=1in,height=1.25in,clip,keepaspectratio]{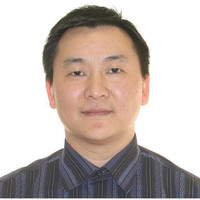}}]{Ye-Kui Wang}
received his BS degree in industrial automation in 1995 from Beijing Institute of Technology, and his PhD degree in information and telecommunication engineering in 2001 from the Graduate School in Beijing, University of Science and Technology of China.
He is currently a Principal Scientist at Bytedance Inc., San Diego, CA, USA. His earlier working experiences and titles include Chief Scientist of Media Coding and Systems at Huawei Technologies, Director of Technical Standards at Qualcomm, Principal Member of Research Staff at Nokia Corporation, etc. His research interests include video coding, storage, transport and multimedia systems.
Dr. Wang has been an active contributor to various multimedia standards, including video codecs, media file formats, RTP payload formats, multimedia streaming protocols and formats, and multimedia application systems, developed by various standardization organizations including ITU-T VCEG, ISO/IEC MPEG, JVT, JCT-VC, JCT-3V, JVET, 3GPP SA4, IETF, AVS, DVB, ATSC, and DECE. He has been chairing the development of OMAF at MPEG, and has been an editor for numerous standards, including VVC, VSEI, OMAF, HEVC, MVC, later versions of AVC, a recent version of ISO base media file format, JPEG AI file format, VVC file format, HEVC file format, layered HEVC file format, SVC file format, a recent version of CMAF, a recent amendment of DASH, ITU-T H.271, RFC 6184, RFC 6190, RFC 7798, RFC 9328, 3GPP TR 26.906, and 3GPP TR 26.948.
\end{IEEEbiography}

\begin{IEEEbiography}[{\includegraphics[width=1in,height=1.25in,clip,keepaspectratio]{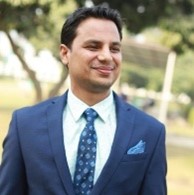}}]
{Shiv Gehlot} received the Ph.D. degree from IIIT-Delhi, India, and is currently a Senior Researcher with Dolby Laboratories, India. His research interests include computer vision, multimodal learning, and generative artificial intelligence.
\end{IEEEbiography}

% \vspace{-4em}
\begin{IEEEbiography}[{\includegraphics[width=1in,height=1.25in,clip,keepaspectratio]{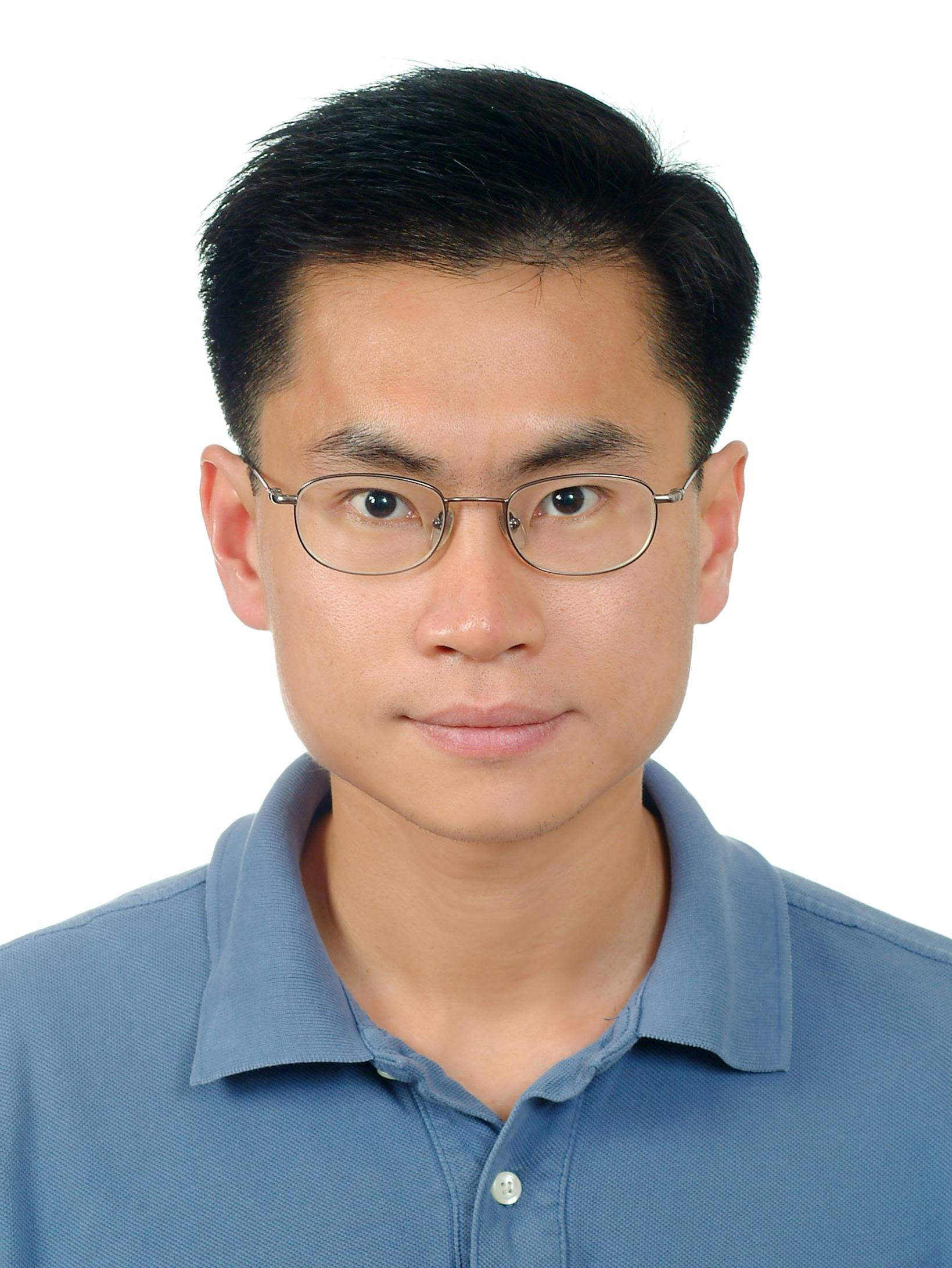}}]
{Guan-Ming Su}(Senior Member, IEEE) obtained his Ph.D. degree from the University of Maryland, College Park. He is currently the Director of Research at Dolby Labs, Sunnyvale, CA, USA. Prior to this, he has been with the R\&D Department, Qualcomm, Inc., San Diego, CA; ESS Technology, Fremont, CA; and Marvell Semiconductor, Inc., Santa Clara, CA. He is the inventor of 210+ U.S./international patents and pending applications. He served as an associate editor in Asia Pacific Signal and Information Processing Association~(APSIPA) Transactions on Signal and Information Processing and IEEE MultiMedia Magazine, and Director of the review board and R-Letter in IEEE Multimedia Communications Technical Committee. He also served in multiple IEEE international conferences such as TPC Co-Chair in ICME 2021, Industry Innovation Forum Chair in ICIP 2023 and 2025, and General Co-Chair in MIPR 2024 and 2025.  He served as VP Industrial Relations and Development in APSIPA 2018--2019. He serves as Vice Chair for Conference in IEEE Technical Committee on Multimedia Computing~(TCMC) since 2021.
\end{IEEEbiography}

\begin{IEEEbiography}[{\includegraphics[width=1in,height=1.25in,clip,keepaspectratio]{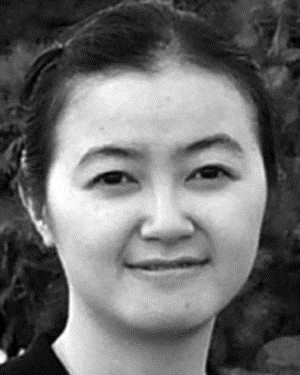}}]
{Peng Yin} received the B.E. degree in electrical engineering from the University of Science and Technology of China in 1996 and the Ph.D. degree in electrical engineering from Princeton University in 2002. She is currently a Director of Imaging Applied Research at Dolby Laboratories, Inc.. She worked at Thomson Inc./Technicolor from 2002 to 2010. She is actively involved in MPEG/VCEG video coding related standardizations. Her research interest is mainly on image/video processing and compression. She received the IEEE Circuits and Systems Society Best Paper Award for her article in the IEEE Transactions on Circuits and Systems for Video Technology in 2003.
\end{IEEEbiography}

\begin{IEEEbiography}[{\includegraphics[width=1in,height=1.25in,clip,keepaspectratio]{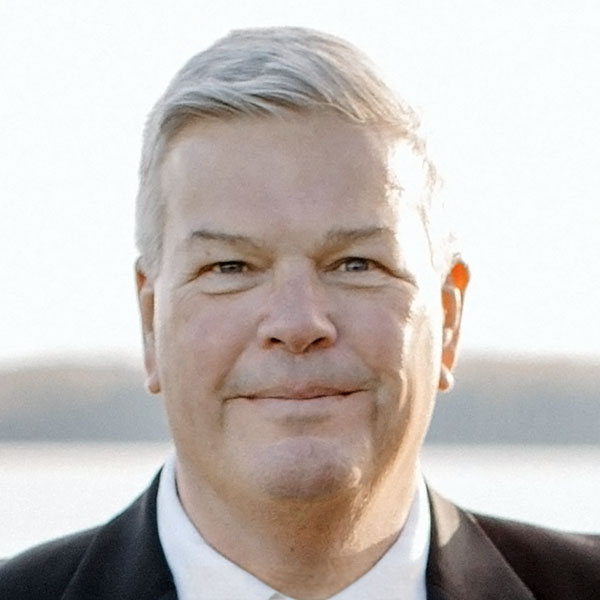}}]
{Sean McCarthy} earned a BS in physics from Rensselaer Polytechnic, and earned his Ph.D. in bioengineering jointly at University of California, Berkeley and University of California, San Francisco. He is currently the Director of Video Strategy and Standards at Dolby Labs, Sunnyvale, CA, USA, where he explores innovations, new use cases, and core technology standards that assist Dolby transform storytelling, and produces new experiences that unleash the potential of entertainment and communications. He brings a unique convergence of expertise in signal processing and the neurobiology of human vision to digital video and entertainment technology. He has published many papers and presented at many industry and technical events in the areas of video processing and applied vision science, and has many issued U.S. and international patents. He is an active contributor, editor, and chair of ad hoc groups for past and ongoing video coding standards developed by ISO/IEC and ITU-T. He is a Fellow of the Society of Motion Picture and Television Engineers and serves on the society’s Board of Editors of the Motion Imaging Journal. Before joining Dolby, he led advances in state-of-the-art of video processing, compression, and practical vision science as a Fellow at ARRIS. He held similar responsibilities as Fellow of Motorola’s technical staff and as Chief Scientist at both Modulus Video and at ViaSense, a spin-off from the University of California Berkeley that developed commercial applications of the human visual system. 
\end{IEEEbiography}

\begin{IEEEbiography}[{\includegraphics[width=1in,height=1.25in,clip,keepaspectratio]{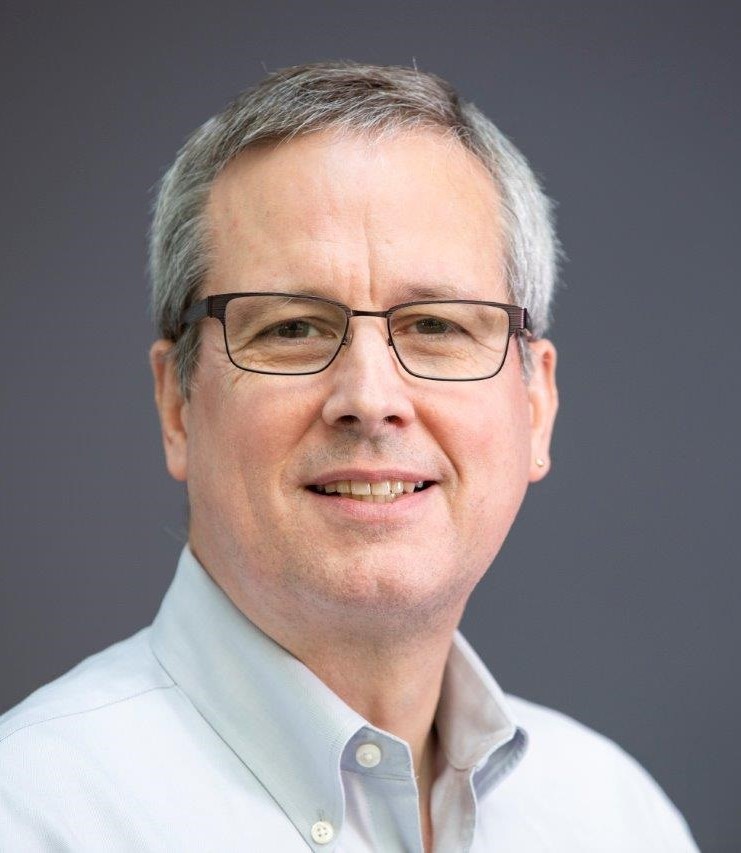}}]
{Gary J. Sullivan} (Fellow, IEEE) received the B.S. and M.Eng. degrees from the University of Louisville in 1982 and 1983, respectively, and the Ph.D. degree from the University of California at Los Angeles, Los Angeles, CA, USA, in 1991. He is currently a Director of Video Research \& Standardization at Dolby Laboratories. He has been a longstanding Chairman/Co-Chairman of various video and image coding standardization activities in ITU-T VCEG, ISO/IEC MPEG, ISO/IEC JPEG, and in their joint collaborative teams since 1996, and since 2021 he has been the chair of ISO/IEC JTC 1 Subcommittee 29, the parent organization of JPEG and MPEG. He led the development of the Advanced Video Coding (AVC) standard (ITU-T H.264 | ISO/IEC 14496-10), the High Efficiency Video Coding (HEVC) standard (ITU-T H.265 | ISO/IEC 23008-2), the Versatile Video Coding (VVC) standard (ITU-T H.266 | ISO/IEC 23090-3), and various other projects. He previously worked for Microsoft Research and was originator and lead designer of the DirectX Video Acceleration (DXVA) video decoding feature of the Microsoft Windows operating system. The team efforts that he has led have been recognized by three Emmy Awards, and he has received the IEEE Masaru Ibuka Consumer Electronics Award and the SMPTE Digital Processing Medal. He is also a fellow of ACM, SMPTE and SPIE.
\end{IEEEbiography}

\end{document}